%% file: main.tex
\documentclass[10pt,twocolumn,letterpaper]{article}

\usepackage[pagenumbers]{cvpr} 
\usepackage[accsupp]{axessibility} 

\input{preamble}

\definecolor{cvprblue}{rgb}{0.21,0.49,0.74}
\usepackage[pagebackref,breaklinks,colorlinks,citecolor=cvprblue]{hyperref}

\def\paperID{****} 
\def\confName{CVPR}
\def\confYear{2024}

\title{TransNeXt: Robust Foveal Visual Perception for Vision Transformers}

\author{Dai Shi\\
{\tt\small daishiresearch@gmail.com}\\
{\tt\small Code: \url{https://github.com/DaiShiResearch/TransNeXt}}
}

\begin{document}

\twocolumn[{
\vspace{-5mm}
\maketitle
\begin{center}
    \vspace{-3mm}
    \includegraphics[width=1.0\textwidth]{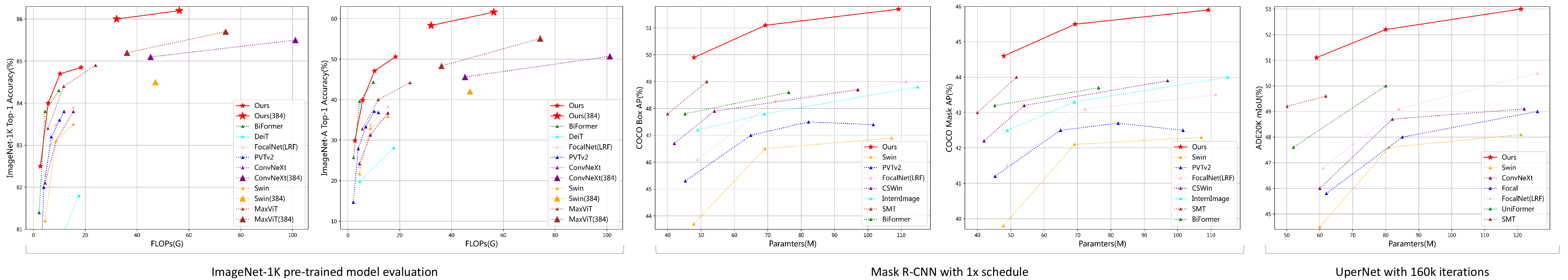}
    \vspace{-6mm}
    \captionof  {figure}{A comprehensive comparison of performance on ImageNet-1K, robustness on ImageNet-A, COCO detection and instance segmentation performance based on Mask R-CNN $1\times$, ADE20K semantic segmentation performance based on UperNet.}
    \vspace{1mm}
	\label{fig:experiment_result}
\end{center}
}]

\input{sec/abstract}
\input{sec/intro}

\input{sec/related}
\input{sec/method}

\input{sec/experiment}

\vspace{-3mm}
\section{Conclusion}
\vspace{-2mm}

In this work, we propose a biomimetic foveal vision design-based token mixer, \textbf{Aggregated Attention}, and a channel mixer with gated channel attention, \textbf{Convolutional GLU}. We combine them to propose a powerful and highly robust visual model, \textbf{TransNeXt}, which achieves state-of-the-art performance in various visual tasks such as classification, detection, and segmentation. The exceptional performance of TransNeXt in multi-scale inference highlights its advantages over large kernel strategies in addressing the issue of depth degradation. Furthermore, we provide a CUDA implementation that achieves up to 103.4\% acceleration in training and 60.5\% acceleration in inference. More detailed experimental data and discussions are included in the appendix.

\appendix
\input{sec/appendix/comparison}

\input{sec/appendix/setting}

\input{sec/appendix/ablation}
\input{sec/appendix/downstream}
\clearpage
\input{sec/appendix/visualization}
\clearpage
\clearpage
{
    \small
    \bibliographystyle{ieeenat_fullname}
    \bibliography{main}
}
\end{document}

%% file: preamble.tex
\usepackage[dvipsnames]{xcolor}

\usepackage{url}
\usepackage{booktabs}       

\usepackage{amsfonts,amsmath}       
\usepackage{nicefrac}       
\usepackage{microtype}      

\usepackage{multirow}       
\usepackage{graphicx}
\usepackage{graphics}
\usepackage{tabularx}
\usepackage{mathrsfs}
\usepackage{makecell}
\usepackage{caption}
\usepackage{wrapfig}
\usepackage{enumitem}
\usepackage{subcaption}

\usepackage{algpseudocode}
\usepackage{amsfonts}       

\usepackage{color, colortbl}
\usepackage{xcolor}
\definecolor{Gray}{gray}{0.95}
\definecolor{orange}{rgb}{0.9,0.5,0}
\definecolor{LightCyan}{rgb}{0.88,1,1}
\usepackage{xfrac}

%% file: sec/abstract.tex
\begin{abstract}
\vspace{-2mm}
Due to the depth degradation effect in residual connections, many efficient Vision Transformers models that rely on stacking layers for information exchange often fail to form sufficient information mixing, leading to unnatural visual perception. To address this issue, in this paper, we propose \textbf{Aggregated Attention}, a biomimetic design-based token mixer that simulates biological foveal vision and continuous eye movement while enabling each token on the feature map to have a global perception. Furthermore, we incorporate learnable tokens that interact with conventional queries and keys, which further diversifies the generation of affinity matrices beyond merely relying on the similarity between queries and keys. Our approach does not rely on stacking for information exchange, thus effectively avoiding depth degradation and achieving natural visual perception.
Additionally, we propose \textbf{Convolutional GLU}, a channel mixer that bridges the gap between GLU and SE mechanism, which empowers each token to have channel attention based on its nearest neighbor image features, enhancing local modeling capability and model robustness. We combine aggregated attention and convolutional GLU to create a new visual backbone called \textbf{TransNeXt}. Extensive experiments demonstrate that our TransNeXt achieves state-of-the-art performance across multiple model sizes. At a resolution of $224^2$, TransNeXt-Tiny attains an ImageNet accuracy of \textbf{84.0\%}, surpassing ConvNeXt-B with \textbf{69\%} fewer parameters. Our TransNeXt-Base achieves an ImageNet accuracy of \textbf{86.2\%} and an ImageNet-A accuracy of \textbf{61.6\%} at a resolution of $384^2$, a COCO object detection mAP of \textbf{57.1}, and an ADE20K semantic segmentation mIoU of \textbf{54.7}.
\end{abstract}

%% file: sec/intro.tex
\begin{figure*}[t]
\centering
    \includegraphics[width=1.\textwidth]{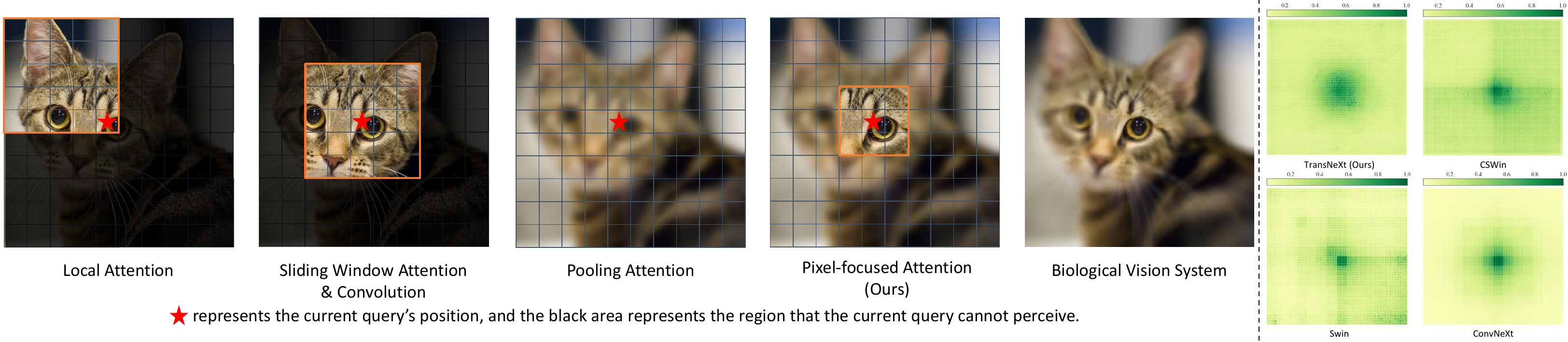}
  \vspace{-2mm}
  \caption{A comparison of prevalent visual information aggregation mechanisms, our proposed method, and biological visual systems (Left) and a visualization comparison of the Effective Receptive Field~\citep{DBLP:conf/nips/LuoLUZ16} between our method and the prevalent backbone networks, using the output at stage 3 (Right). Each ERF image is generated by averaging over 5000 $224^2$-sized images from ImageNet-1K validation set.}
  \vspace{-5mm}
  \label{fig:biological_vision}
\end{figure*}
\vspace{-3mm}
\section{Introduction}
\vspace{-2mm}
\label{Sec:intro}

The Vision Transformer (ViT)~\citep{DBLP:conf/iclr/DosovitskiyB0WZ21} has emerged as a popular backbone architecture for various computer vision tasks in recent years. The ViT model comprises two key components: the self-attention layer (token mixer) and the MLP layer (channel mixer). The self-attention mechanism plays a crucial role in feature extraction by dynamically generating an affinity matrix through similarity computations between queries and keys. This global information aggregation method has demonstrated remarkable feature extraction potential, with no inductive bias like convolution~\citep{lecun1995convolutional}, and can build powerful data-driven models. However, the transformer encoder design of vision transformers, originally developed for language modeling~\citep{vaswani2017attention}, exhibits inherent limitations in downstream computer vision tasks. Specifically, the computation of the global affinity matrix in self-attention poses a challenge due to its quadratic complexity and high memory consumption, which restricts its application on high-resolution image features. 

In order to mitigate the computational and memory burdens imposed by the quadratic complexity inherent in the self-attention mechanism, a plethora of sparse attention mechanisms have been proposed in previous studies. One such representative method is local attention~\citep{DBLP:conf/cvpr/Liu0LYXWN000WG22}, which restricts attention within a window on the feature map. However, due to the limited receptive field, this method often requires alternating stacking with different types of token mixers to achieve cross-window information exchange. Another representative method spatially downsamples the keys and values of attention (such as pooling~\citep{DBLP:conf/iccv/WangX0FSLL0021,DBLP:journals/corr/abs-2106-13797,DBLP:conf/iccv/WuXCLDY021}, grid sampling~\citep{DBLP:conf/eccv/TuTZYMBL22}). This method, due to its sacrifice of the query’s fine-grained perception of the feature map, also has certain limitations. Recent studies~\citep{DBLP:conf/nips/ChuTWZRWXS21,DBLP:conf/eccv/TuTZYMBL22} have alternately stacked spatial downsampling attention and local attention, achieving commendable performance results.

However, recent studies~\citep{DBLP:conf/nips/DeS20,DBLP:conf/nips/VeitWB16} and experiments~\citep{DBLP:journals/prl/KimCJLJK23} have shown that deep networks with residual blocks~\citep{DBLP:conf/cvpr/HeZRS16} behave like ensembles of shallower networks, indicating that the cross-layer information exchange achieved by stacking blocks may not be as effective as anticipated.

On the other hand, both local attention and spatial downsampling attention differ significantly from the workings of biological vision. Biological vision possesses higher acuity for features around the visual focus and lower acuity for distant features. Moreover, as the eyeball moves, this characteristic of biological vision remains consistent for pixels at any position in the image, implying pixel-wise translational equivariance. However, in local attention based on window partitioning, tokens at the window edge and center are not treated equivalently, presenting a clear discrepancy.

We have observed that due to depth degradation effects, many efficient ViT models are unable to form sufficient information mixing through stacking. Even with a deep stack of layers, the traces of their window partitioning always form unnatural artifacts, as shown in Fig~\ref{fig:biological_vision}. To address this issue, we investigate a visual modeling approach that closely aligns with biological vision to mitigates potential model depth degradation and achieve information perception closer to human foveal vision. To this end, we initially introduce \textbf{Pixel-focused Attention}, which employs a dual-path design. In one path, each query has fine-grained attention to its nearest neighbor features, while in the other path, each query has coarse-grained attention to spatial downsampled features, allowing for a global perception. This approach operates on a per-pixel basis, effectively simulating the continuous movement of the eyeball.  Furthermore, we incorporate query embedding and positional attention mechanisms into pixel-focused attention, leading to the proposal of \textbf{Aggregated Pixel-focused Attention}, which we abbreviate as \textbf{Aggregated Attention}. This approach further diversifies the generation of affinity matrices beyond merely relying on the similarity between queries and keys, thereby achieving the aggregation of multiple attention mechanisms within a single attention layer. We also reevaluate the design requirements of the channel mixer in vision transformers and propose a novel channel mixer named \textbf{Convolutional GLU}. This mixer is more apt for image tasks and integrates local feature-based channel attention to enhance model robustness.

We introduce \textbf{TransNeXt}, a hierarchical visual backbone network that incorporates \textbf{aggregated attention} as a token mixer and \textbf{convolutional GLU} as a channel mixer. Through comprehensive evaluation across image classification, object detection, and segmentation tasks, we demonstrate the efficacy of these mixing components. Our TransNeXt-Tiny, pretrained solely on ImageNet-1K, achieves an ImageNet accuracy of \textbf{84.0\%}, surpassing ConvNeXt-B. In COCO object detection, it attains a box mAP of \textbf{55.1} using the DINO detection head, outperforming ConvNeXt-L pretrained at a resolution of  $384^2$ by 1.7. Our TransNeXt-Small/Base, fine-tuned at a resolution of $384^2$ for merely \textbf{5 epochs}, achieves an ImageNet accuracy of \textbf{86.0\%/86.2\%}, surpassing the previous state-of-the-art MaxViT-Base fine-tuned for 30 epochs by 0.3\%/0.5\%. Moreover, when evaluated on the highly challenging ImageNet-A test set at a resolution of $384^2$, our TransNeXt-Small/Base models achieve an impressive top-1 accuracy of \textbf{58.3\%/61.6\%}, significantly outperforming ConvNeXt-L by 7.6\%/10.9\%, setting a new benchmark of robustness for ImageNet-1K supervised models.

In summary, our contributions are as follows: 
\begin{enumerate}[leftmargin=*]
    \item Proposing \textbf{pixel-focused attention}, a token mixer closely aligns with biological foveal vision and mitigates potential model depth degradation. This novel attention mechanism works on a per-pixel basis, effectively simulating the continuous movement of the eyeball and highly aligning with the focal perception mode of biological vision. It possesses visual priors comparable to convolution.  
    
    \item Proposing \textbf{aggregated attention}, an enhanced version of pixel-focused attention, which further aggregates two types of non-QKV attention mechanisms into pixel-focused attention. Notably, we propose a highly efficient approach within this framework, with the additional computational overhead accounting for a mere 0.2\%-0.3\% of the entire model, leading to a exceptionally cost-effective unification of QKV attention, LKV attention, and QLV attention within a single mixer layer.
    
    \item Proposing \textbf{length-scaled cosine attention} that enhances the extrapolation capability of existing attention mechanisms for multi-scale input. This allows TransNeXt to achieve superior large-scale image extrapolation performance compared to pure convolutional networks.
    \item Proposing \textbf{convolutional GLU}, which incorporates channel attention based on nearest neighbor image features. In comparison to convolutional feed-forward, it realizes the attentionalization of the channel mixer with fewer FLOPs, thereby effectively enhancing the model’s robustness.
    \item Introducing \textbf{TransNeXt}, a visual backbone that delivers state-of-the-art performance in various visual tasks such as image classification, object detection, and semantic segmentation among models of similar size. It also exhibits state-of-the-art robustness.
\end{enumerate}

\vspace{-3mm}

%% file: sec/related.tex
\section{Related Work}
\label{Sec:RelatedWork}
\vspace{-2mm}

\textbf{Vision transformers}: Vision Transformer (ViT)~\citep{DBLP:conf/iclr/DosovitskiyB0WZ21} was the first to introduce transformer architecture to visual tasks, where images are segmented into non-overlapping patches and subsequently linearly projected into token sequences, which are later encoded by a transformer encoder. When trained with large-scale pretraining data or thoughtfully designed training strategies, ViT models outperform convolutional neural networks (CNNs)\citep{lecun1995convolutional,DBLP:journals/cacm/KrizhevskySH17,DBLP:conf/cvpr/HeZRS16}, exhibiting remarkable performance in image classification and other downstream tasks. 

\textbf{Non-QKV attention variants}: In self-attention, the dynamic affinity matrix is generated through the interaction between queries and keys. Recently, several studies~\citep{DBLP:journals/corr/abs-2005-00743,DBLP:conf/cvpr/Li0WLSZZC21,DBLP:journals/pami/YuanHJFY23,DBLP:conf/cvpr/ArarSB22} have explored the use of learnable tokens as a replacement for the original queries or keys to generate dynamic affinity matrices. Involution~\citep{DBLP:conf/cvpr/Li0WLSZZC21} and VOLO~\citep{DBLP:journals/pami/YuanHJFY23}, for instance, use learnable tokens to replace the original keys, resulting in dynamic affinity matrices that are exclusively correlated with queries. In contrast, QnA~\citep{DBLP:conf/cvpr/ArarSB22} utilizes learnable tokens to replace queries, leading to dynamic affinity matrices that are only correlated with keys. Both methods have shown effectiveness. 

\textbf{Biomimetic vision modeling}: Human vision exhibits higher acuity for features around the visual focus and lower acuity for distant features. This biomimetic design has been integrated into several machine vision models~\citep{DBLP:conf/nips/MinZLC22,DBLP:conf/nips/YangLDG22,DBLP:journals/corr/abs-2107-00641}. Specifically, Focal Transformer~\citep{DBLP:journals/corr/abs-2107-00641} designs a visual attention based on this concept, but it operates based on window partitioning. Tokens located at the window edges cannot obtain natural foveal vision, and its window-wise manner cannot simulate the continuous movement of the human eyeball. Our approach effectively addresses these shortcomings.
\vspace{-5mm}

%% file: sec/method.tex
\section{Method}
\vspace{-2mm}
\label{Sec:Method}
\subsection{Aggregated Pixel-focused Attention}
\vspace{-2mm}

\begin{figure*}[t]
\centering
    \includegraphics[width=1.\textwidth]{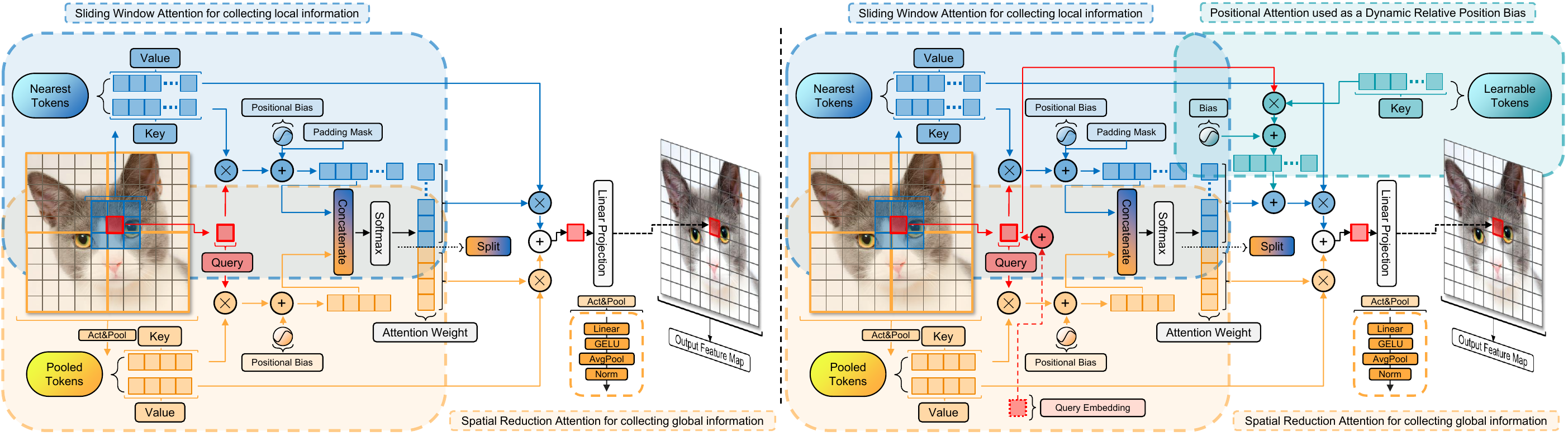}
  \vspace{-5mm}
  \caption{An illustration of the comparison between pixel-focused attention (left) and aggregated attention (right). Both have a feature size of 10$\times$10, a window size of 3$\times$3, and a pool size of 2$\times$2.}
  \label{fig:aggregated_attention}
  \vspace{-5mm}
\end{figure*}

\subsubsection{Pixel-focused Attention}
\vspace{-2mm}

Inspired by the functioning of biological visual systems, we have designed a pixel-focused attention mechanism that possesses fine-grained perception in the vicinity of each query, while concurrently maintaining a coarse-grained awareness of global information. To achieve the pixel-wise translational equivariance inherent in eyeball movements, we employ a dual-path design incorporating query-centered sliding window attention and pooling attention. Furthermore, to induce coupling between the two attention paths, we compute the importance in the same softmax for the query-key similarity results of both paths. This results in a competition between fine-grained and coarse-grained features, transforming pixel-focused attention into a multi-scale attention mechanism.

Given an input $X \in \mathbb{R}^{C \times H \times W}$, we now focus on the operations performed on a single pixel in the input feature map. We define a set of pixels within a sliding window centered at pixel at $(i,j)$ as $\rho(i,j)$. For a fixed window size of $k\times k$, $\Vert\rho(i,j)\Vert=k^2$. Concurrently, we define the set of pixels obtained from pooling the feature map as $\sigma(X)$. Given a pooling size of $H_p \times W_p$, $\Vert\sigma(X)\Vert=H_pW_p$. Therefore, \textbf{p}ixel-\textbf{f}ocused \textbf{a}ttention (\textbf{PFA}) can be described as follows:
\begin{equation}
    \begin{aligned}
    &S_{(i,j)\sim \rho(i,j)}=Q_{(i,j)}K_{\rho(i,j)}^T\\
    &S_{(i,j)\sim\sigma(X)}=Q_{(i,j)}K_{\sigma(X)}^T
    \end{aligned}
\vspace{-2mm}
\label{eq:similarity}
\end{equation}

\begin{equation}
\begin{split} 
A_{(i,j)}=&\mathrm{softmax}\\
&(\frac{\mathrm{Concat}(S_{(i,j)\sim\rho(i,j)},S_{(i,j)\sim\sigma(X)})}{\sqrt{d}} +B_{(i,j)}) 
\end{split}
\vspace{-2mm}
\label{eq:attn_weight}
\end{equation}

\begin{equation}
\begin{aligned}
A_{(i,j)\sim\rho(i,j)},A_{(i,j)\sim\sigma(X)}=&\mathrm{Split}(A_{(i,j)})\\&\text{with size }[k^2,H_pW_p]
\end{aligned}
\vspace{-2mm}
\label{eq:attn_weight_spilt}
\end{equation}

\begin{equation}
\textbf{PFA}(X_{(i,j)})=A_{(i,j)\sim\rho(i,j)}V_{\rho(i,j)}+A_{(i,j)\sim\sigma(X)}V_{\sigma(X)}
\label{eq:pfa_output}
\end{equation}

\textbf{Activate and Pool}: In order to utilize the linear complexity mode of PFA for large-scale image inference in subsequent applications, we employ parameter-free adaptive average pooling for downsampling in the spatial dimension. However, the average pooling operator significantly loses information. Therefore, we use a single-layer neural network for projection and activation before feature map pooling to compress and extract useful information in advance, thereby improving the information compression rate after downsampling. After pooling, we once again use layer normalization to normalize the output to ensure the variance consistency of $X$ and $\sigma(X)$. The downsampling operator we propose, termed ‘Activate and Pool’, can be expressed by the following equation:

\begin{equation}
\sigma(X)=\mathrm{LayerNorm}(\mathrm{AvgPool}(\mathrm{GELU}(\mathrm{Linear}(X))))
\label{eq:act_and_pool}
\end{equation}

We replaced the downsampling module in PVTv2-li~\citep{DBLP:journals/corr/abs-2106-13797} with our ‘activate and pool’ mechanism and designed a 2M-sized model for ablation experiments on CIFAR-100~\citep{krizhevsky2009learning}. Our module improved the top-1 accuracy of PVTv2-li from 68.1\% to 70.4\%, demonstrating the effectiveness of this approach.

\textbf{Padding mask}: In the sliding window path, pixels located at the edge of the feature map inevitably compute similarities with zero-padding outside the boundary. To prevent these zero similarities from influencing the softmax operation, we employ a padding mask to set these results to $-\infty$.

\subsubsection{Aggregating Diverse Attentions in a Single Mixer}\label{sec:lkv_qlv_attention}
\vspace{-2mm}

\textbf{Query embedding}: Several vision-language models~\citep{DBLP:conf/icml/0001LXH22,DBLP:conf/icml/0008LSH23} utilize queries originating from the textual modality to perform cross-attention on keys derived from the visual modality, thereby achieving cross-modal information aggregation to complete Visual Question Answering (VQA) tasks. Moreover, it has been proven effective and efficient to incorporate and optimize learnable prefix query tokens when fine-tuning these multimodal models to adapt to specific subtasks.

A natural extension of this idea is to incorporate these learnable query tokens into the attention mechanism of the backbone network for well-defined tasks such as image classification, object detection, and semantic segmentation, and directly optimize them. This approach has been validated by previous work~\citep{DBLP:conf/cvpr/ArarSB22} for its effectiveness.

This method differs from traditional QKV attention as it does not use queries from the input but learns a query defined by the current task to perform cross-attention. Therefore, we categorize this method as \textbf{L}earnable-\textbf{K}ey-\textbf{V}alue (\textbf{LKV}) attention, drawing a parallel to QKV attention. We found that adding a learnable \textbf{Q}uery \textbf{E}mbedding (\textbf{QE}) to all query tokens in traditional QKV attention can achieve similar information aggregation effects with negligible additional overhead. We only need to modify Equation~\ref{eq:similarity} as follows:
\begin{equation}
    \begin{aligned}
    &S_{(i,j)\sim \rho(i,j)}=(Q_{(i,j)}+\text{QE})K_{\rho(i,j)}^T
    \\
    &S_{(i,j)\sim\sigma(X)}=(Q_{(i,j)}+\text{QE})K_{\sigma(X)}^T
    \end{aligned}
\label{eq:query_embedding}
\end{equation}

\textbf{Positional attention}: An alternative approach to information aggregation is the use of a set of learnable keys that interact with queries originating from the input to obtain attention weights, \textit{i.e.}, \textbf{Q}uery-\textbf{L}earnable-\textbf{V}alue (\textbf{QLV}) attention. This method differs from traditional QKV attention as it disrupts the one-to-one correspondence between keys and values, resulting in learning more implicit relative positional information for the current query. Consequently, it is often employed in conjunction with a sliding window in visual tasks~\citep{DBLP:conf/cvpr/Li0WLSZZC21,DBLP:journals/pami/YuanHJFY23}. Unlike static affinity matrices such as convolution or relative position bias, the affinity matrix generated in this way takes into account the impact of the current query and can dynamically adapt based on it.  We have observed that this data-driven modeling approach exhibits greater robustness compared to static relative position bias and can further enhance locality modeling capabilities. Leveraging this feature, we introduce a set of learnable tokens $T \in \mathbb{R}^{d \times k^2}$ in each attention head, allowing these tokens to interact with queries to obtain additional dynamic position bias and add it to $A_{(i,j)\sim\rho(i,j)}$. Using this enhancement only requires an additional computational overhead of $HWk^2C$. We only need to modify Equation~\ref{eq:pfa_output} as follows:

\begin{equation}
\begin{aligned}
\textbf{PFA}(X_{(i,j)})=&(A_{(i,j)\sim\rho(i,j)}+Q_{(i,j)}T)V_{\rho(i,j)}\\&+A_{(i,j)\sim\sigma(X)}V_{\sigma(X)}
\end{aligned}
\label{eq:positional_attention}
\end{equation}

\subsubsection{Overcoming Multi-scale Image Input}
\vspace{-2mm}
\textbf{Length-scaled cosine attention}: In contrast to the scaled dot product attention, the scaled cosine attention, which employs cosine similarity, has been observed to generate more moderate attention weights~\citep{DBLP:conf/emnlp/HenryDPC20,DBLP:conf/cvpr/Liu0LYXWN000WG22} and effectively enhance the training stability of large visual models~\citep{DBLP:conf/icml/0001DMPHGSCGAJB23}. The scaled cosine attention typically multiplies an additional learnable coefficient $\lambda$ to the cosine similarity results of queries and keys, enabling the attention mechanism to effectively ignore insignificant tokens~\citep{DBLP:conf/emnlp/HenryDPC20}. Recent studies~\citep{DBLP:journals/tacl/Hahn20,DBLP:conf/acl/0001C22} have discovered that as the length of the input sequence increases, the confidence of the attention output decreases. Therefore, the scaling factor of the attention mechanism should be related to the length of the input sequence~\citep{DBLP:conf/acl/0001C22}. ~\cite{kexuefm-8823} further proposed that the design of attention should exhibit entropy invariance to facilitate better generalization to unknown lengths. ~\cite{kexuefm-8823} provided an estimate of the entropy of the scaled dot product attention with a sequence length $n$ when queries and keys are approximated as vectors with a magnitude of $\sqrt{d}$:
\begin{equation}
\mathcal{H}_i \approx \log n - 0.24\lambda d + \mathscr{O}(1)
\label{eq:entropy_invariance}
\end{equation}

For cosine similarity, we define the queries and keys with $\ell_2$-normalization applied along their head dimensions as $\hat Q$ and $\hat K$ respectively, both of which have magnitudes of 1. To maintain entropy invariance and disregard constant terms, we set $\lambda \approx \frac{\log n}{0.24}$. Given that Equation~\ref{eq:entropy_invariance} is merely an estimate, we set $\lambda = \tau\log n$, where $\tau$ is a learnable variable initialized to $\frac{1}{0.24}$ for each attention head. We propose \textbf{length-scaled cosine attention} as follows:
\begin{figure*}[t]
\centering
    \includegraphics[width=0.95\textwidth]{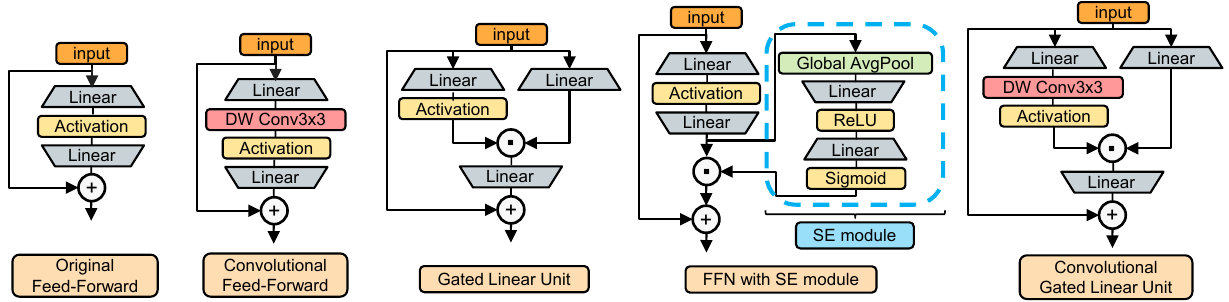}
    \vspace{-2mm}
  \caption{Comparison of prevalent  channel mixer designs and Convolutional GLU}
  \label{feedforward_variants}
  \vspace{-5mm}
\end{figure*}
\begin{figure*}[t]
\centering
    \includegraphics[width=1.\textwidth]{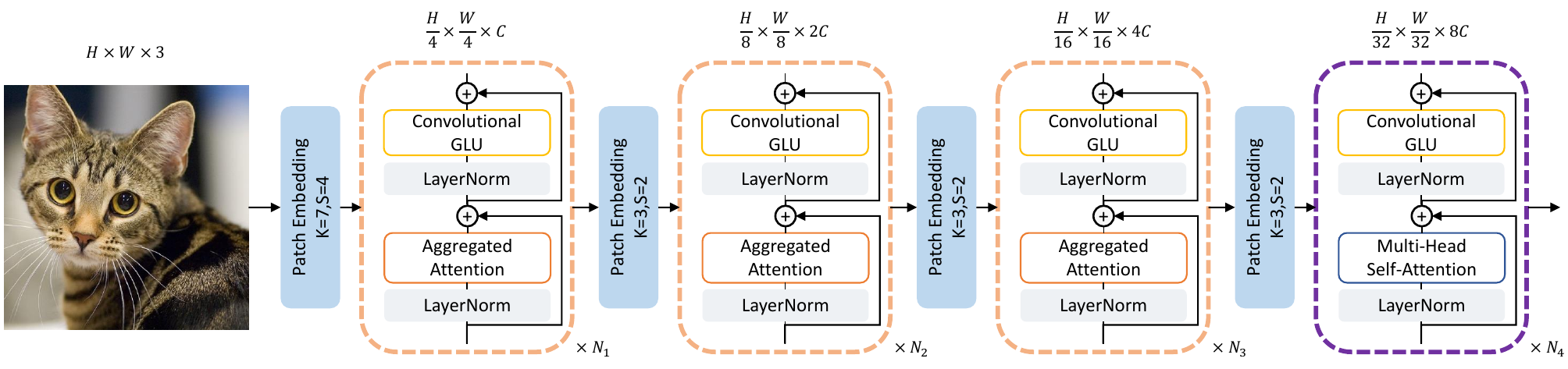}
  \vspace{-7mm}
  \caption{An illustration of TrasnNeXt architecture.}
  \label{fig:model_architecture}
  \vspace{-7mm}
\end{figure*}
\begin{equation}
\mathrm{Attention}(Q,K,V)=\mathrm{softmax}(\tau\log N*\hat Q\hat K^T)V
\label{eq:length_scaled_cosine_attention}
\end{equation}

Here, $N$ denotes the count of effective keys each query interacts with, excluding the count of masked tokens. Specifically, when applied in a transformer decoder~\citep{vaswani2017attention}, future tokens masked by a causal mask should not be counted in $N$. In the context of pixel-focused attention, $N$ is calculated as 
$N_{(i,j)}=\Vert\rho(i,j)\Vert+\Vert\sigma(X)\Vert-\Vert\mu(i,j)\Vert$
, where $\mu(i,j)$ represents the set of padding-masked tokens at position $(i,j)$.

\textbf{Position bias}: 
To further enhance the extrapolation capability of pixel-focused attention for multi-scale image inputs, we employ different methods to calculate $B_{(i,j)\sim\rho(i,j)}$ and $B_{(i,j)\sim\sigma(X)}$ on two paths. On the pooling feature path, we use \textbf{log}-spaced \textbf{c}ontinuous \textbf{p}osition \textbf{b}ias (\textbf{log-CPB})~\citep{DBLP:conf/cvpr/Liu0LYXWN000WG22}, a 2-layer MLP with a ReLU~\citep{DBLP:conf/icml/NairH10} to compute $B_{(i,j)\sim\sigma(X)}$ from the spatial relative coordinates $\Delta_{(i,j)\sim\sigma(X)}$ between $Q_{(i,j)}$ and $K_{\sigma(X)}$. On the sliding window path, we directly use a learnable $B_{(i,j)\sim\rho(i,j)}$. On one hand, this is because the size of the sliding window is fixed and does not require extrapolation of unknown relative position biases through log-CPB, thus saving computational resources. On the other hand, we observe that using log-CPB to calculate $B_{(i,j)\sim\rho(i,j)}$ results in performance degradation. We believe this is because $\Delta_{(i,j)\sim\sigma(X)}$ represents the spatial relative coordinates between fine-grained tokens and coarse-grained tokens, while $\Delta_{(i,j)\sim\rho(i,j)}$represents the spatial relative coordinates between fine-grained tokens, and their numerical meanings are different. We discuss these details further in appendix.

\textbf{Aggregated attention}: By applying the aforementioned diverse attention aggregation methods and techniques for enhancing the extrapolation capability for multi-scale inputs, we propose an enhanced version of pixel-focused attention, termed aggregated pixel-focused attention, which we abbreviate as \textbf{A}ggregated \textbf{A}ttention (\textbf{AA}). It can be described as follows:
\begin{equation}
    \begin{aligned}
    &S_{(i,j)\sim \rho(i,j)}=(\hat Q_{(i,j)}+\text{QE})\hat K_{\rho(i,j)}^T\\
    &S_{(i,j)\sim\sigma(X)}=(\hat Q_{(i,j)}+\text{QE})\hat K_{\sigma(X)}^T
    \end{aligned}
\vspace{-2mm}
\label{eq:aa_similarity}
\end{equation}

\begin{equation}
\begin{aligned}
B_{(i,j)}=\mathrm{Concat}(B_{(i,j)\sim\rho(i,j)},\textbf{log-CPB}(\Delta_{(i,j)\sim\sigma(X)}))
\end{aligned}
\vspace{-2mm}
\label{eq:aa_rpe}
\end{equation}

\begin{equation}
\begin{aligned}
A_{(i,j)}=&\mathrm{softmax}(\tau\log N*
\\&\mathrm{Concat}(S_{(i,j)\sim\rho(i,j)},S_{(i,j)\sim\sigma(X)})+B_{(i,j)}) 
\end{aligned}
\vspace{-2mm}
\label{eq:aa_attn_weight}
\end{equation}

\begin{equation}
\begin{aligned}
A_{(i,j)\sim\rho(i,j)},A_{(i,j)\sim\sigma(X)}=&\mathrm{Split}(A_{(i,j)})\\
&\text{ with size }[k^2,H_pW_p]
\end{aligned}
\vspace{-2mm}
\label{eq:aa_attn_weight_spilt}
\end{equation}

\begin{equation}
\begin{aligned}
\textbf{AA}(X_{(i,j)})=&(A_{(i,j)\sim\rho(i,j)}+\hat Q_{(i,j)}T)V_{\rho(i,j)}\\
&+A_{(i,j)\sim\sigma(X)}V_{\sigma(X)}
\end{aligned}
\label{eq:aa_output}
\end{equation}

\vspace{-3mm}
\subsubsection{Feature Analysis}
\vspace{-2mm}
\textbf{Computational complexity}: Given an input $X \in \mathbb{R}^{C \times H \times W}$, a pooling size of $H_p \times W_p$, and a window size of $k \times k$, we consider the impact of ‘activate and pool’ operation and linear projection. The computational complexities of pixel-focused attention and aggregated attention are:
\begin{equation}
\begin{aligned}
\Omega(\textbf{PFA})=&5HWC^2+2H_pWpC^2\\&+2HWH_pW_pC+2HWk^2C
\end{aligned}
\vspace{-2mm}
\label{eq:pfa_complexity}
\end{equation}

\begin{equation}
\begin{aligned}
\Omega(\textbf{AA})&=\Omega(\textbf{PFA})+HWk^2C\\
&=5HWC^2+2H_pWpC^2\\&\quad+2HWH_pW_pC+3HWk^2C
\end{aligned}
\label{eq:aa_complexity}
\end{equation}

We observe that when the pooling size $H_p \times W_p$ is set to a value independent of the input size, Both $\Omega(\textbf{PFA})$ and $\Omega(\textbf{AA})$ scales linearly with the length of the input sequence. This implies that both PFA and AA can perform inference in a \textbf{linear complexity mode}.

\textbf{Optimal accuracy-efficiency trade-off}: Through empirical studies, we observed that the size of the sliding window has a negligible impact on model performance. Consequently, we employed the minimal form of a $3\times 3$ sliding window to capture features near the visual focus, significantly reducing computational and memory consumption. We attribute this to the presence of pooling feature paths, which endow each query with a global receptive field, thereby greatly diminishing the need to expand the sliding window size to extend the receptive field. Detailed ablation study results and discussions can be found in appendix.

\subsection{Convolutional GLU}
\vspace{-2mm}
\subsubsection{Motivation}
\vspace{-2mm}

\textbf{Gated channel attention in ViT era}: Previous work, represented by the Squeeze-and-Excitation (SE) mechanism~\citep{DBLP:journals/pami/HuSASW20}, first introduced channel attention into the field of computer vision, which uses a branch with an activation function to gate the network output. In gated channel attention, the gating branch has more decision-making power than the value branch, and it ultimately determines whether the corresponding output elements are zeroed. From this perspective, the SE mechanism cleverly uses features after global average pooling as the input of the gating branch, achieving a largest receptive field for better decision-making and solving the problem of insufficient receptive field in CNNs structures at the same time. However, in the ViT era, global receptive fields are no longer scarce. Various global token mixers represented by self-attention have achieved higher quality global information aggregation than global average pooling. This makes the global pooling method used by the SE mechanism show some shortcomings, such as this method makes all tokens on the feature map share the same gating signal, making its channel attention lack flexibility and too coarse-grained. Despite this, it’s worth noting that ViT structures lack channel attention. Recent research~\citep{DBLP:conf/icml/ZhouYXXAFA22} has found that incorporating the SE mechanism into a channel mixer can effectively enhance model robustness, as shown in Fig.~\ref{feedforward_variants}.

\textbf{Convolution in ViT era}: Recent studies~\citep{chu2021conditional,DBLP:conf/iclr/IslamJB20} have shown that introducing a $3\times3$ depthwise convolution~\citep{DBLP:conf/cvpr/Chollet17} into the vision transformer can be viewed as a form of conditional position encoding (CPE)~\citep{chu2021conditional}, which effectively captures positional information from zero-padding.

\vspace{-5mm}
\subsubsection{Rethinking Channel Mixer Design}
\vspace{-3mm}
The Gated Linear Unit (GLU)~\citep{DBLP:conf/icml/DauphinFAG17,DBLP:journals/corr/abs-2002-05202} is a channel mixer that has been shown to outperform Multi-Layer Perceptron (MLP) in various natural language processing tasks. GLU consists of two linear projections that are element-wise multiplied, with one projection being activated by a gating function. Unlike the SE mechanism, its gating signal for each token is derived from the token itself and does not have a larger receptive field than the value branch.

\textbf{More elegant design}: We found that simply adding a minimal form of $3\times3$ depthwise convolution before the activation function of GLU’s gating branch can make its structure conform to the design concept of gated channel attention and convert it into a gated channel attention mechanism based on nearest neighbor features. We named this method \textbf{Convolutional GLU}, as shown in Fig.~\ref{feedforward_variants}.

\textbf{Feature analysis}: Each token in \textbf{Conv}olutional \textbf{GLU} (\textbf{ConvGLU}) possesses a unique gating signal, based on its nearest fine-grained features. This addresses the overly coarse-grained drawback of the global average pooling in the SE mechanism. It also meets the needs of some ViT models without position encoding design that require position information provided by depthwise convolution. Moreover, the value branch of this design still maintains the same depth as MLP and GLU, making it backpropagation-friendly. When keeping the parameter volume consistent with the Convolutional Feed-Forward (ConvFFN)~\citep{DBLP:journals/corr/abs-2106-13797} with an expansion ratio of $R$ and a convolution kernel size of $k\times k$, the computational complexity of ConvGLU is $2RHWC^2+\frac{2}{3}RHWCk^2$, which is less than the $2RHWC^2+RHWCk^2$ of ConvFFN. These attributes render ConvGLU a simple yet more robust mixer, satisfying the diverse requirements of ViTs.

\vspace{-2mm}
\subsection{Architecture Design of TransNeXt}
\vspace{-2mm}

In order to ensure consistency in subsequent ablation experiments~\ref{sec:roadmap_PVT}, TransNeXt adopts the same four-stage hierarchical backbone and overlapping patch embedding as PVTv2~\citep{DBLP:journals/corr/abs-2106-13797}. The pooling feature size of the aggregated attention in stages 1-3 is also set to $\frac{H}{32}\times\frac{W}{32}$, identical to PVTv2. In stage 4, as the feature map size has been reduced to $\frac{H}{32}\times\frac{W}{32}$, the feature pooling module cannot function properly. We employ a modified version of multi-head self-ttention (MHSA) that applies query embedding and length-scaled cosine attention. This is consistent with PVTv2’s use of MHSA in the fourth stage. For the channel mixer in stages 1-4, we use convolutional GLU with GELU~\citep{hendrycks2016gaussian} activation. The expansion ratio also follows PVTv2’s [8,8,4,4] setting. To ensure consistency with typical MLP parameters, the hidden dimension of convolutional GLU is $\frac{2}{3}\times$ of the set value. Furthermore, we set the head dimension to be 24 for divisibility by 3 in the channel dimension. The specific configurations of TransNeXt variants can be found in appendix.

%% file: sec/experiment.tex
\vspace{-2mm}
\section{Experiment}
\vspace{-5mm}
\label{Sec:Experiment}

\begin{table}[htpb]
\begin{center}
  \resizebox{1\linewidth}{!}{
    \begin{tabular}{l|cccccccc}
    \toprule[1.5pt]
     Model & \makecell{\#Params. \\ (M)} & \makecell{FLOPs \\ (G)} & \makecell{IN-1K $\uparrow$ \\Top-1(\%)}& \makecell{IN-C $\downarrow$ \\mCE(\%)} & \makecell{IN-A $\uparrow$ \\Top-1(\%)}  & \makecell{IN-R $\uparrow$ \\Top-1(\%)}  & \makecell{Sketch $\uparrow$ \\Top-1(\%)}  & \makecell{IN-V2 $\uparrow$ \\Top-1(\%)}  \\
    \midrule[1.5pt]
    \multicolumn{9}{c}{\textbf{ImageNet-1K 224$^2$ pre-trained models}}\\
    PVT-Tiny~\citep{DBLP:conf/iccv/WangX0FSLL0021} & 13.2 & 1.9 & 75.1 & 79.6 & 8.2 & 33.7 & 21.3 & 63.0\\
    PVTv2-B1~\citep{DBLP:journals/corr/abs-2106-13797} & 14.0 & 2.1 & 78.7 & 62.6 & 14.7 & 41.8 & 28.9 & 66.9\\
    BiFormer-T~\citep{DBLP:journals/corr/abs-2303-08810} & 13.1 & 2.2 & 81.4 & 55.7 & 25.7 & 45.4 & 31.5 & 70.6\\
    EfficientFormerv2-S2~\citep{DBLP:journals/corr/abs-2212-08059} & 12.7 & 1.3 & 81.6 &\textbf{--}& \textbf{--}& \textbf{--}& \textbf{--}& \textbf{--}\\
    \rowcolor{LightCyan}
    \textbf{TransNeXt-Micro (Ours)} & 12.8 & 2.7 & \textbf{82.5} & \textbf{50.8} & \textbf{29.9} & \textbf{45.8} & \textbf{33.0} & \textbf{72.6}\\
    \midrule
    DeiT-Small/16~\citep{DBLP:conf/icml/TouvronCDMSJ21} & 22.1 & 4.6 & 79.9 & 54.6 & 19.8 & 41.9 & 29.1 & 68.4\\
    Swin-T~\citep{DBLP:conf/iccv/LiuL00W0LG21} & 28.3 & 4.5 & 81.2 & 62.0 & 21.7 & 41.3 & 29.0 & 69.7\\
    PVTv2-B2~\citep{DBLP:journals/corr/abs-2106-13797} & 25.4 & 4.0 & 82.0 & 52.6 & 27.9 & 45.1 & 32.8 & 71.6\\
    ConvNeXt-T~\citep{DBLP:conf/cvpr/0003MWFDX22} & 28.6 & 4.5 & 82.1 & 53.2 & 24.2 & 47.2 &33.8 &71.0\\
    Focal-T~\citep{DBLP:journals/corr/abs-2107-00641} & 29.1 & 4.9 & 82.2 & \textbf{--}& \textbf{--}& \textbf{--}& \textbf{--}& \textbf{--}\\
    FocalNet-T (LRF)~\citep{DBLP:conf/nips/YangLDG22} & 28.6 & 4.5 & 82.3 &55.0 & 23.5 & 45.1 & 31.8 & 71.2\\
    MaxViT-Tiny~\citep{DBLP:conf/eccv/TuTZYMBL22} & 30.9 & 5.6 & 83.4 & 49.6 & 32.8 & 48.3 & 36.3 & 72.9\\
    BiFormer-S~\citep{DBLP:journals/corr/abs-2303-08810} & 25.5 & 4.5 & 83.8 & 48.5 & 39.5 & 49.6 & 36.4 & 73.7\\
    \rowcolor{LightCyan}
    \textbf{TransNeXt-Tiny (Ours)} & 28.2 & 5.7 & \textbf{84.0} & \textbf{46.5} & \textbf{39.9} & \textbf{49.6} & \textbf{37.6} & \textbf{73.8}\\
    \midrule
    Swin-S~\citep{DBLP:conf/iccv/LiuL00W0LG21} & 49.6 & 8.7 & 83.1 & 54.9 & 32.9 & 44.9 & 32.0 & 72.1\\
    ConvNeXt-S~\citep{DBLP:conf/cvpr/0003MWFDX22} & 50.2 & 8.7 & 83.1 & 49.5 & 31.3& 49.6& 37.1&72.5\\
    PVTv2-B3~\citep{DBLP:journals/corr/abs-2106-13797} & 45.2 & 6.9 & 83.2 & 48.0 & 33.3 & 49.2 & 36.7 & 73.0\\
    Focal-S~\citep{DBLP:journals/corr/abs-2107-00641} & 51.1 & 9.1 & 83.5 & \textbf{--}& \textbf{--}& \textbf{--}& \textbf{--}& \textbf{--}\\
    FocalNet-S (LRF)~\citep{DBLP:conf/nips/YangLDG22} & 50.3 & 8.7 & 83.5 & 51.0 & 33.8 & 47.7 & 35.1 & 72.7\\
    PVTv2-B4~\citep{DBLP:journals/corr/abs-2106-13797} & 62.6 & 10.1 & 83.6 & 46.5 & 37.1 & 49.8 & 37.5 & 73.5\\
    BiFormer-B~\citep{DBLP:journals/corr/abs-2303-08810} & 56.8 & 9.8 & 84.3 & 47.2 & 44.3 & 49.7 & 35.3 & 74.0\\
    MaxViT-Small~\citep{DBLP:conf/eccv/TuTZYMBL22} & 68.9 & 11.7 & 84.4 & 46.4 & 40.0 & 50.6 & 38.3 & 74.0\\
    \rowcolor{LightCyan}
    \textbf{TransNeXt-Small (Ours)}  & 49.7 & 10.3 & \textbf{84.7} & \textbf{43.9} & \textbf{47.1} & \textbf{52.5} & \textbf{39.7} & \textbf{74.8}\\
    \midrule
    DeiT-Base/16~\citep{DBLP:conf/icml/TouvronCDMSJ21} & 86.6 & 17.6 & 81.8 & 48.5 & 28.1 & 44.7 & 32.0 & 70.9\\
    Swin-B~\citep{DBLP:conf/iccv/LiuL00W0LG21} & 87.8 & 15.4 & 83.5 & 54.5 & 35.9 & 46.6 & 32.4 & 72.3\\
    PVTv2-B5~\citep{DBLP:journals/corr/abs-2106-13797} & 82.0 & 11.8 & 83.8 & 45.9 & 36.8 & 49.8 & 37.2 & 73.4\\
    Focal-B~\citep{DBLP:journals/corr/abs-2107-00641} & 89.8 & 16.0 & 83.8 & \textbf{--}& \textbf{--}& \textbf{--}& \textbf{--}& \textbf{--}\\
    ConvNeXt-B~\citep{DBLP:conf/cvpr/0003MWFDX22} & 88.6 & 15.4 & 83.8 & 46.8 &36.7 &51.3 & 38.2&73.7\\
    FocalNet-B (LRF)~\citep{DBLP:conf/nips/YangLDG22} & 88.7 & 15.4 & 83.9 & 49.5 & 38.3 & 48.1 & 35.7 & 73.5\\
    \rowcolor{LightCyan}
    \textbf{TransNeXt-Base (Ours)} & 89.7 & 18.4 & \textbf{84.8} & \textbf{43.5} & \textbf{50.6} & \textbf{53.9} & \textbf{41.4} & \textbf{75.1}\\
    \midrule
    MaxViT-Base~\citep{DBLP:conf/eccv/TuTZYMBL22} & 119.5 & 24.0 & \textbf{84.9} & 43.6 & 44.2 & 52.5 & 40.1 & 74.5\\
    \midrule[1.5pt]
    \multicolumn{9}{c}{\textbf{ImageNet-1K 384$^2$ fine-tuned models}}\\
    Swin-B~\citep{DBLP:conf/iccv/LiuL00W0LG21} & 87.8 & 47.1 &84.5&\textbf{--}&42.0&47.2&33.4&73.2\\
    ConvNeXt-B~\citep{DBLP:conf/cvpr/0003MWFDX22}& 88.6&45.2&85.1 &\textbf{--}&45.6&52.9&39.5&75.2\\
    MaxViT-Small~\citep{DBLP:conf/eccv/TuTZYMBL22} & 68.9 & 36.1 & 85.2&\textbf{--}&48.3&\textbf{--}&\textbf{--}&\textbf{--} \\
    ConvNeXt-L~\citep{DBLP:conf/cvpr/0003MWFDX22}& 197.8&101.1&85.5&\textbf{--}&50.7&54.6&41.0&76.0\\
    MaxViT-Base~\citep{DBLP:conf/eccv/TuTZYMBL22} & 119.5 & 74.2 & 85.7&\textbf{--}&55.1&\textbf{--}&\textbf{--}&\textbf{--} \\
    \rowcolor{LightCyan}
    \textbf{TransNeXt-Small (Ours)} & 49.7 & 32.1 &\textbf{86.0} &\textbf{--}&\textbf{58.3}&\textbf{56.4}&\textbf{43.2}&\textbf{76.8}\\
    \rowcolor{LightCyan}
    \textbf{TransNeXt-Base (Ours)} & 89.7 & 56.3 &\textbf{86.2} &\textbf{--}& \textbf{61.6}&\textbf{57.7}&\textbf{44.7}&\textbf{77.0}\\
    \bottomrule[1.5pt]
    \end{tabular}
    }
    \end{center}
    \vspace{-7mm}
    \caption{A comprehensive comparison on the ImageNet-1K classification and additional robustness test sets.}
    \vspace{-4mm}
    \label{tab:image_classification_robustness}
\end{table}

\textbf{ImageNet-1K classification}: Our code is implemented based on PVTv2~\citep{DBLP:journals/corr/abs-2106-13797} and follows the DeiT~\citep{DBLP:conf/icml/TouvronCDMSJ21} recipe for training. The model is trained from scratch on the ImageNet-1K~\citep{DBLP:conf/cvpr/DengDSLL009} dataset for 300 epochs, leveraging automatic mixed precision (AMP) across $8\times$ GPUs. The specific hyperparameters employed during training are detailed in appendix. To conduct a comprehensive evaluation of the model’s robustness, we utilize several additional test sets. These include ImageNet-C~\citep{DBLP:conf/iclr/HendrycksD19}, a $224^2$-sized test set that applies algorithmic distortions to ImageNet-1K validation set; ImageNet-A~\citep{DBLP:conf/cvpr/HendrycksZBSS21}, a test set comprising adversarial examples; ImageNet-R~\citep{DBLP:conf/iclr/HendrycksD19}, an extended test set containing samples that ResNet-50~\citep{DBLP:conf/cvpr/HeZRS16} failed to classify correctly; ImageNet-Sketch~\citep{DBLP:conf/nips/WangGLX19}, which contains hand-drawn images; and ImageNet-V2~\citep{DBLP:conf/icml/RechtRSS19}, an extended test set that employs the same sampling strategy as ImageNet-1K.

\textbf{Experimental results}: The experimental results, presented in Table~\ref{tab:image_classification_robustness}, establish that our proposed model sets a new benchmark in ImageNet-1K accuracy and robustness across various scales. Specifically, our TransNeXt-Micro model achieves a top-1 accuracy of \textbf{82.5\%} on ImageNet-1K, surpassing the FocalNet-T(LRF) while utilizing 55\% fewer parameters. Similarly, our TransNeXt-Tiny model achieves a top-1 accuracy of \textbf{84.0\%}, outperforming ConvNeXt-B with a reduction of 69\% in parameters. Remarkably, at a resolution of $384^2$, our TransNeXt-Small/Base model surpasses the larger MaxViT-Base model by \textbf{0.3\%/0.5\%} respectively after only \textbf{5 epochs} of fine-tuning, compared to the 30 epochs used by MaxViT-Base. In terms of robustness, our model exhibits superior performance on five additional test sets. Notably, on the most challenging ImageNet-A test set, TransNeXt demonstrates a significant advantage in robustness as the model scales up. On ImageNet-A at a resolution of $224^2$, our TransNeXt-Base surpasses MaxViT-Base by 6.4\%. At a resolution of $384^2$, our TransNeXt-Small/Base achieves an impressive ImageNet-A accuracy of \textbf{58.3\%/61.6\%}, significantly outperforming ConvNeXt-L by 7.6\%/10.9\%, while their parameter counts are only 25\% and 45\% of ConvNeXt-L, respectively.

\textbf{Object detection and instance segmentation}: We employed a Mask R-CNN~\citep{DBLP:journals/pami/HeGDG20} detection head, trained under a $1\times$ schedule, to evaluate the performance of ImageNet-1K pretrained TransNeXt on object detection and instance segmentation on the COCO~\citep{DBLP:conf/eccv/LinMBHPRDZ14} dataset. The experimental results are presented in Fig~\ref{fig:experiment_result}. Our model demonstrated comprehensive superiority when compared with previous state-of-the-art models. Notably, even our tiny model surpassed the base models of FocalNet, InternImage and CSWin in terms of $AP^b$. Similarly, we utilized a DINO~\citep{DBLP:conf/iclr/0097LL000NS23} detection head, also trained under a $1\times$ schedule, to further assess the potential of our model for object detection. Our TransNeXt-Tiny model achieved an $AP^b$ of 55.1 under a 4-scales setting, surpassing ConvNeXt-L($AP^b$ of 53.4 in 4-scales setting) 1.7 with only 14\% of the latter’s backbone parameters. Our TransNeXt-Base achieved an $AP^b$ of 57.1 under a 5-scales setting, approaching the performance of Swin-L($AP^b$ of 57.2 in 5-scales setting) pretrained on ImageNet-22K.

\textbf{Semantic segmentation}: We used UperNet~\citep{DBLP:conf/eccv/XiaoLZJS18} and Mask2Former~\citep{DBLP:conf/cvpr/ChengMSKG22} methods to train the ImageNet-1K pretrained TransNeXt at a resolution of $512^2$ for 160k iterations, and evaluated its semantic segmentation performance on ADE20K~\citep{DBLP:journals/ijcv/ZhouZPXFBT19}. Under the UperNet method, as shown in Fig~\ref{fig:experiment_result}, our TransNeXt demonstrated comprehensive superiority over previous methods across all sizes. Our TransNeXt-Base even surpassed ConvNeXt-B (mIoU 52.6), which was pretrained on ImageNet-22K and further trained at a resolution of $640^2$. Similarly, under the Mask2Former method, our TransNeXt-Small achieved an mIoU of 54.1, surpassing Swin-B (mIoU 53.9) which was pretrained on ImageNet-22K and further trained at a resolution of $640^2$. Furthermore, our TransNeXt-Base achieved an mIoU of 54.7. These results indicate that our method has the potential to transcend model size limitations and break through data volume barriers.

Our model demonstrates a even more pronounced performance advantage in dense prediction tasks compared to classification tasks. We believe this validates the effectiveness of the biomimetic vision design of aggregated attention, which enables a more natural visual perception at earlier stages compared to previous methods, as depicted in Fig~\ref{fig:biological_vision}.

\vspace{-2mm}
\subsection{Multi-scale Inference}\label{sec:multiscale_inference}
\vspace{-2mm}
\begin{figure}[t]
\centering
    \includegraphics[width=0.485\textwidth]{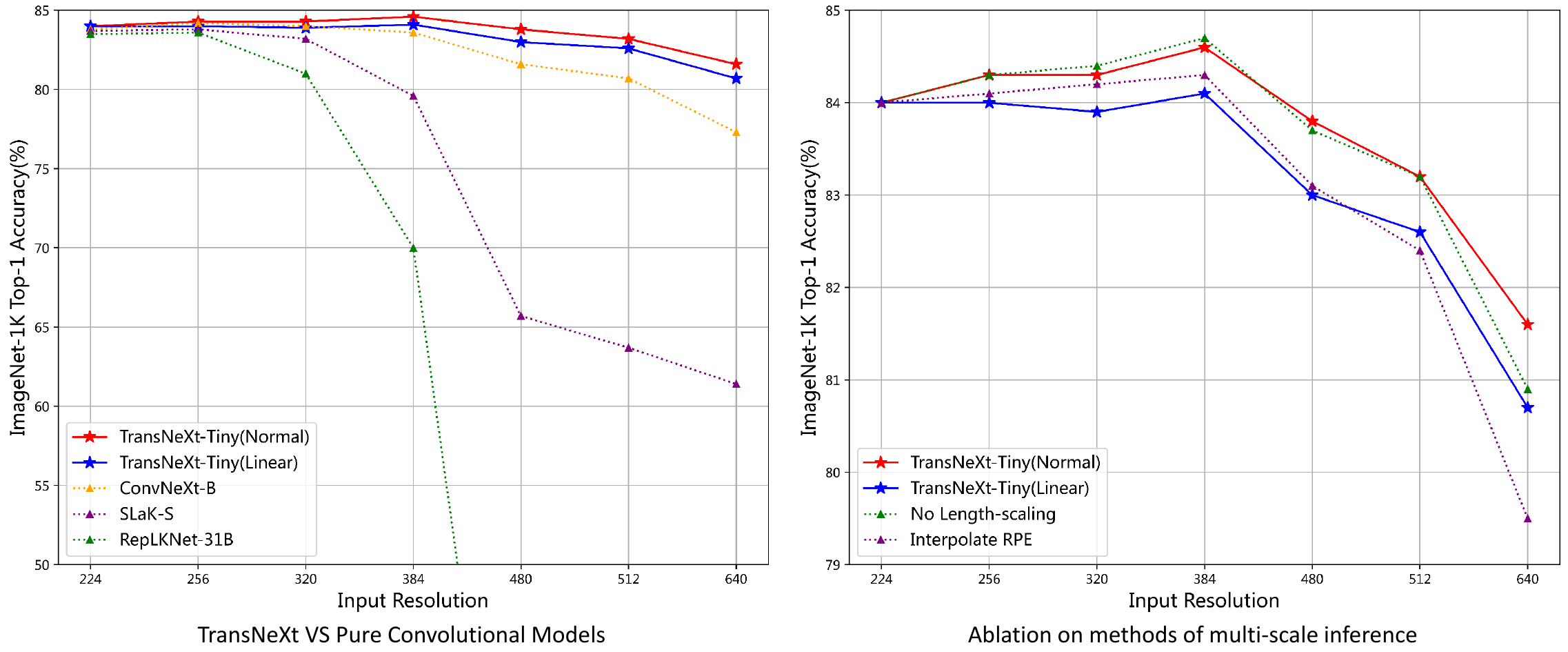}
  \vspace{-7mm}
  \caption{The left figure shows the comparison results of TransNeXt-Tiny under normal and linear inference modes with the pure convolution models on multi-scale image inference performance, while the right figure shows the impact of our positional encoding design and length-scaled cosine attention on this aspect.}
  \label{fig:multi_scale_inference}
  \vspace{-6mm}
\end{figure}
During inference, TransNeXt in normal mode sets $H_p$ and $W_p$ to $\frac{1}{32}$ of the input size, while in linear mode, these are fixed at $7\times7$. As depicted in Fig~\ref{fig:multi_scale_inference} (left), TransNeXt outperforms pure convolutional solutions in both normal and linear modes. Large convolutional kernel schemes~\citep{DBLP:conf/cvpr/Ding0HD22,DBLP:conf/iclr/LiuCCCXWKPMW23}, also proposed to address depth degradation, exhibit significant performance decline during large image size inference. This reveals the advantage of our approach over large kernel schemes in addressing this issue. For instance, RepLKNet-31B only achieves 0.9\% accuracy at a resolution of $640^2$. In traditional opinions, pure convolutional models have better multi-scale applicability than ViT models, and such experimental results imply that this opinion needs to be re-examined. The performance decline of large kernel strategies also merits further investigation by the research community.

Fig~\ref{fig:multi_scale_inference} (right) illustrates the impact of length-scaled cosine and the use of interpolation for position bias on performance. Length-scaling becomes significant at a resolution of $640^2$, indicating that sequence length variations exceeding 8× in softmax start to notably diminish the confidence of scaled cosine attention. The application of interpolation for relative position biases results in a substantial performance decline, emphasizing the effectiveness of using extrapolative positional encoding (log-CPB) in multi-scale inference.

\vspace{-2mm}
\subsection{A roadmap from PVT to TransNeXt}\label{sec:roadmap_PVT}
\vspace{-4mm}

\begin{table}[htpb]
\begin{minipage}{1.0\linewidth}
\footnotesize
\centering
\setlength{\tabcolsep}{1.8pt}
\resizebox{1\linewidth}{!}{
    \begin{tabular}{c|c|cccccccc}
    \toprule[1.5pt]
     Step& \makecell{Method} & \makecell{\#Params. \\ (M)} & \makecell{FLOPs \\ (G)} & \makecell{IN-1K $\uparrow$ \\Top-1(\%)}& \makecell{IN-C $\downarrow$ \\mCE(\%)} & \makecell{IN-A $\uparrow$ \\Top-1(\%)}  & \makecell{IN-R $\uparrow$ \\Top-1(\%)}  & \makecell{Sketch $\uparrow$ \\Top-1(\%)}  & \makecell{IN-V2 $\uparrow$ \\Top-1(\%)}  \\
    \midrule
    0 & PVT-Tiny~\citep{DBLP:conf/iccv/WangX0FSLL0021} & 13.2 & 1.9 & 75.1 & 79.6 & 8.2 & 33.7 & 21.3 & 63.0\\
    1 & PVTv2-B1~\citep{DBLP:journals/corr/abs-2106-13797} & 14.0 & 2.1 & 78.7 \tiny{\textcolor{red}{(+3.6)}} & 62.6 \tiny{\textcolor{red}{(+17.0)}} 
    & 14.7 \tiny{\textcolor{red}{(+6.5)}} & 41.8 \tiny{\textcolor{red}{(+8.1)}} & 28.9 \tiny{\textcolor{red}{(+7.6)}} & 66.9 \tiny{\textcolor{red}{(+3.9)}} \\
    2 & Deeper and Thinner & 14.9 & 2.3 & 80.08 \tiny{\textcolor{red}{(+1.38)}} & 55.3 \tiny{\textcolor{red}{(+7.3)}} & 19.7 \tiny{\textcolor{red}{(+5.0)}} & 43.2 \tiny{\textcolor{red}{(+1.4)}} & 31.1 \tiny{\textcolor{red}{(+2.2)}} & 69.2 \tiny{\textcolor{red}{(+2.3)}} \\
    3 & + More Heads & 14.9 & 2.3 & 80.12 \tiny{\textcolor{red}{(+0.04)}} & 55.0 \tiny{\textcolor{red}{(+0.3)}} & 19.2 \tiny{\textcolor{blue}{(-0.5)}} & 43.5 \tiny{\textcolor{red}{(+0.3)}} & 31.5 \tiny{\textcolor{red}{(+0.4)}} & 69.4 \tiny{\textcolor{red}{(+0.2)}} \\
    4 & ConvFFN$\rightarrow$GLU & 14.8 & 2.2 & 79.7 \tiny{\textcolor{blue}{(-0.42)}} & 59.5 \tiny{\textcolor{blue}{(-4.5)}} & 18.9 \tiny{\textcolor{blue}{(-0.3)}} & 39.3 \tiny{\textcolor{blue}{(-4.2)}} & 26.8 \tiny{\textcolor{blue}{(-4.7)}} & 69.0 \tiny{\textcolor{blue}{(-0.4)}} \\
    5 & GLU$\rightarrow$ConvGLU & 14.9 & 2.2 & 80.9 \tiny{\textcolor{red}{(+1.2)}} & 54.6 \tiny{\textcolor{red}{(+4.9)}} & 23.5 \tiny{\textcolor{red}{(+4.6)}} & 44.3 \tiny{\textcolor{red}{(+5.0)}} & 32.7 \tiny{\textcolor{red}{(+5.9)}} & 70.6 \tiny{\textcolor{red}{(+1.6)}} \\
    6 & SRA$\rightarrow$PFA & 12.8 & 2.7 & 81.8 \tiny{\textcolor{red}{(+0.9)}} & 51.7 \tiny{\textcolor{red}{(+2.9)}} & 26.9 \tiny{\textcolor{red}{(+3.4)}} & 45.2 \tiny{\textcolor{red}{(+0.9)}} & 33.3 \tiny{\textcolor{red}{(+0.6)}} & 71.6 \tiny{\textcolor{red}{(+1.0)}} \\
    7 & + Positional Attention & 12.8 & 2.7 & 82.2 \tiny{\textcolor{red}{(+0.4)}} & \textbf{50.7} \tiny{\textcolor{red}{(+1.0)}} & \textbf{31.0} \tiny{\textcolor{red}{(+4.1)}} & \textbf{46.4} \tiny{\textcolor{red}{(+1.2)}} & \textbf{34.1} \tiny{\textcolor{red}{(+0.8)}} & 72.0 \tiny{\textcolor{red}{(+0.4)}} \\
    \rowcolor{LightCyan}
    8 & + Query Embedding & 12.8 & 2.7 & \textbf{82.5} \tiny{\textcolor{red}{(+0.3)}} & 50.8 \tiny{\textcolor{blue}{(-0.1)}} & 29.9 \tiny{\textcolor{blue}{(-1.1)}} & 45.8 \tiny{\textcolor{blue}{(-0.6)}} & 33.0 \tiny{\textcolor{blue}{(-1.1)}} & \textbf{72.6} \tiny{\textcolor{red}{(+0.6)}} \\

     \bottomrule[1.5pt]
    \end{tabular}
    }
    \captionsetup{font=footnotesize}  
    \vspace{-3mm}
    \caption{The ablation experiments demonstrate the full roadmap from PVT-Tiny to TransNeXt-Micro. In step 1, PVTv2 introduces Overlapping Patch Embedding and Convolutional Feed-Forward (ConvFFN). In step 2, we made PVTv2 consistent with TransNeXt-Tiny in terms of height and width, with a head dimension of 48. In step 3, we decreased the head dimension to 24 and increased the number of attention heads.}
    \label{tab:ablation}
    \vspace{-4mm}
\end{minipage}
\end{table}

\textbf{Effectiveness of our method}: The efficacy of our proposed convolutional GLU (ConvGLU) , pixel-focused attention, positional attention, and query embedding is demonstrated through ablation experiments from step 4 to 8. In the stages of step 4 to 5, step 6, and step 7 to 8, we replaced convolutional feed-forward (ConvFFN) with ConvGLU, spatial-reduction attention (SRA) with pixel-focused attention (PFA), and pixel-focused attention with aggregated attention, respectively. These three substitutions resulted in accuracy improvements of 0.8\%, 0.9\%, and 0.7\% on ImageNet-1K, and 4.3\%, 3.4\%, and 3.0\% on the ImageNet-A test set, respectively, indicating the significant contribution of these three components to performance. It is noteworthy that the introduction of QLV and LKV mechanisms in pixel-focused attention required only an additional 0.2\% parameters (from 12.78M to 12.81M) and 0.3\% computational overhead (from 2.65G to 2.66G), yet the performance improvement was significant, thereby achieving a cost-effective trade-off. Moreover, in step 4, replacing ConvFFN with GLU led to a significant performance decline, underscoring the necessity of the $3\times3$ depthwise convolution~\citep{DBLP:conf/cvpr/Chollet17} as conditional position encodings (CPE)~\citep{chu2021conditional}, particularly as PVTv2's SRA~\citep{DBLP:journals/corr/abs-2106-13797} did not use any other positional encoding at this stage. Therefore, step 5 also demonstrated the effectiveness of using ConvGLU as positional encoding.

%% file: sec/appendix/comparison.tex
\section{Equivalent Form of Pixel-Focused Attention}

Mathematically, pixel-focused attention is equivalent to the following form:
\begin{equation}
    \begin{aligned}
    &K_{concat}=Concat(K_{\rho(i,j)},K_{\sigma(X)})\\
    &V_{concat}=Concat(V_{\rho(i,j)},V_{\sigma(X)})
    \end{aligned}
\label{eq:kv_concat}
\end{equation}

\begin{equation}
\begin{split} 
\textbf{PFA}(X_{(i,j)})=&\mathrm{softmax}(\frac{Q_{(i,j)}K_{concat}^T}{\sqrt{d}} +B_{(i,j)})V_{concat} 
\end{split}
\label{eq:pfa_output_concat}
\end{equation}

This form is more concise mathematically. However, in parallel computation, merging $\rho(i,j)$ and $\sigma(X)$ on key and value to form $K_{concat}, V_{concat}~in~   \mathbb{R}^{nh\times HW\times(H_p W_p+k^2)\times d}$ creates two large temporary tensors. This results in significant memory usage and memory access pressure, severely slowing down the model speed. Therefore, in practical applications, we use the method of separately calculating the attention of the two paths and only adding the results. Calculating the attention weight from the concatenated similarity result in the same softmax is crucial, as it ensures the mathematical equivalence of these two forms.

\begin{figure*}[t]
\centering
    \includegraphics[width=1.\textwidth]{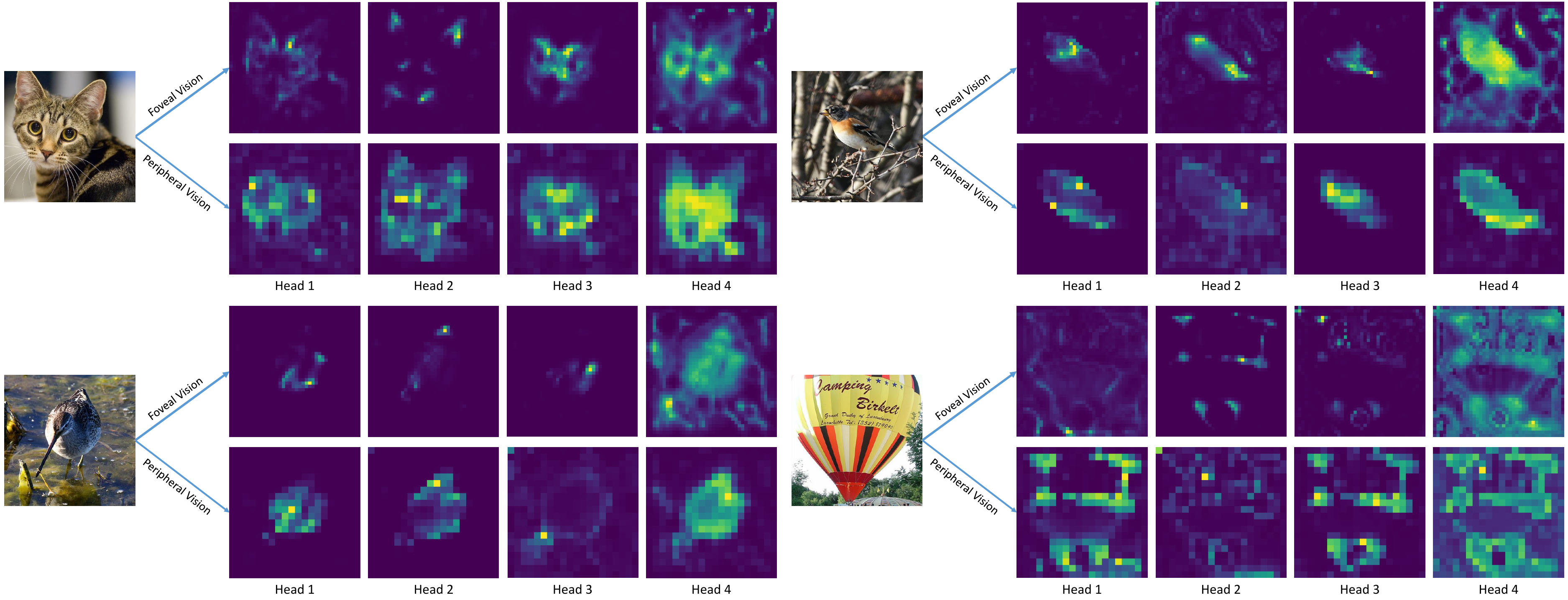}
  \vspace{-6mm}
  \caption{The attention map of foveal and peripheral vision when the visual focus is centered. The central query token of the feature map is utilized to compute $\text{softmax}(\tau\log N*\hat Q_{[center]} \hat K^T)$ and $\text{softmax}(\tau\log N*\hat Q_{[center]} \hat K_{\sigma(X)}^T)$. For effective visualization, we employ a high-resolution image input of $640^2$ and calculate the attention map using the final attention layer of stage 3. It's important to note that during the model's standard operation, the foveal vision perception only utilizes the features of the $k\times k$ area near the query. However, given that this area is too small to provide sufficient information for observation, we use undownsampled global features for visualization purposes, allowing us to discern the features of interest to the foveal vision perception.}
  \label{fig:attention_visualization}
\end{figure*}

\section{Comparative Analysis of Human Vision and Attention Visualization}

The human visual system is characterized by a dichotomy between foveal and peripheral vision. The foveal vision, covering only 1 to 2 degrees of the central field of view, is optimized for high sensitivity, while the peripheral vision, with a much larger receptive field, is optimized for a broad field of view. This dichotomy suggests that humans primarily utilize peripheral vision for object localization and navigation, but lack precision in detail. To compensate for this, the human eye executes rapid movements, known as saccades, allowing for the processing of information from multiple fields of view and the subsequent integration of this information. Recent research~\citep{noauthor_2022-bg} on biological vision suggests that when superior foveal vision information is available, the human brain does not simply discard peripheral vision information, but compares the inputs from both, weighing their relative reliability to complete the integration of information. This process is a highly complex computation known as transsaccadic perception.

Pixel-focused attention, in comparison to human vision, maintains a high degree of similarity with the human visual system in numerous design aspects. Empirical studies~\ref{sec:ablation_window_size} demonstrate that the sliding window path in pixel-focused attention, which simulates foveal vision and continuous eye movements, can achieve superior performance with a minimal $3\times3$ window size for perception. This experimental result aligns closely with the extremely narrow field of view coverage of human foveal vision. Particularly in the context of high-resolution image input, the coverage range of the $3\times3$ sliding window is exceedingly small. The pooling path in pixel-focused attention, which simulates human peripheral vision, maintains full-image perception and overlaps with the fine-grained perception area of the sliding window path. The features of these two window paths compete and correlate in the same softmax. This design enables the model to compare the detailed features perceived by foveal vision with the comprehensive rough outline of the object perceived by peripheral vision in terms of similarity, thereby integrating and calibrating information across the field of view. From the perspective of biological vision, the current design of pixel-focused attention can effectively simulate the highly complex transsaccadic perception in human visual perception. In contrast, the Focal Transformer, which employs a window partitioning approach, is unable to simulate the continuous movements of the human eye. Furthermore, its method of pooling only the peripheral windows prevents the model from fully perceiving the comprehensive rough outline of the object, thereby increasing its inconsistency with human transsaccadic perception.

In the terminal attention layer of TransNeXt’s third stage, we employ the query token located at the center position $(H\div2,W\div2)$ of the feature map (subsequently represented by the $[center]$ subscript), to calculate $\text{softmax}(\tau\log N*\hat Q_{[center]}\hat K^T)$, and $\text{softmax}(\tau\log N*\hat Q_{[center]}\hat K_{\sigma(X)}^T)$ respectively. This allows us to visualize the attention map of the image feature perception in the sliding window path and the pooling feature path at the central position of the picture, with the results presented in Fig~\ref{fig:attention_visualization}. It can be observed that in the same head, the sliding window path simulating foveal vision consistently maintains interest in specific textures, fluff, and high sharpness edges, while the pooling path simulating peripheral vision consistently has a reliable grasp of the rough outline of the object. The features of the two paths are further matched and calibrated in the same softmax, resulting in more accurate output. We posit that this working method closely resembles the integration method of foveal and peripheral vision found in human vision research, further substantiating the high consistency of our biomimetic design with human vision.

%% file: sec/appendix/setting.tex
\section{Detailed Settings}

\subsection{Configurations of TransNeXt Variants} 
\begin{table*}[htpb]
\begin{center}
  \resizebox{1\linewidth}{!}{
  \begin{tabular}{l|c|c|c|c|c|c|c}
    \toprule[1.5pt]
     Model & Channels & Head dims& Blocks & MLP ratio & Token mixer & Window size & Pool size\\
    \midrule
    TransNeXt-Micro & [48, 96, 192, 384] & 24 &[2, 2, 15, 2] & [8, 8, 4, 4] & \textbf{A-A-A-M} & [3, 3, 3, \textbf{--}] & [7, 7, 7,  \textbf{--}]\\
    TransNeXt-Tiny & [72, 144, 288, 576] & 24 & [2, 2, 15, 2] & [8, 8, 4, 4] & \textbf{A-A-A-M} & [3, 3, 3,  \textbf{--}] & [7, 7, 7,  \textbf{--}]\\
    TransNeXt-Small & [72, 144, 288, 576] & 24 & [5, 5, 22, 5] & [8, 8, 4, 4] & \textbf{A-A-A-M} & [3, 3, 3,  \textbf{--}] & [7, 7, 7,  \textbf{--}]\\
    TransNeXt-Base & [96, 192, 384, 768] & 24 & [5, 5, 23, 5] & [8, 8, 4, 4] & \textbf{A-A-A-M} & [3, 3, 3, \textbf{--}] & [7, 7, 7,  \textbf{--}] \\
    \bottomrule[1.5pt]
    \end{tabular}
    }
\end{center}
\vspace{-6mm}
\caption{The configurations of TransNeXt variants. The value of pool size  is calculated at $224^2$ resolution. \textbf{A} = aggregated attention, while \textbf{M} = multi-head self-attention.}
\label{table:transnext_config}
\end{table*}

\subsection{Training Settings for ImageNet-1K} 
To ensure reproducibility and consistency with prior work, we adopt the training strategy of PVTv2~\citep{DBLP:journals/corr/abs-2106-13797}, which incorporates various data augmentation techniques, including Random Augmentation~\citep{DBLP:conf/cvpr/CubukZSL20}, Mixup~\citep{DBLP:conf/iclr/ZhangCDL18}, CutMix~\citep{DBLP:conf/iccv/YunHCOYC19}, and Random Erasing~\citep{DBLP:conf/aaai/Zhong0KL020}. To regularize our model, we employ Label Smoothing~\citep{DBLP:conf/cvpr/SzegedyVISW16} and DropPath~\citep{DBLP:conf/eccv/HuangSLSW16}. We optimize our model using AdamW~\citep{DBLP:conf/iclr/LoshchilovH19} optimizer with a gradient clipping norm of 1.0 and a weight decay of 0.05. The initial learning rate for all models is set to $10^{-3}$, with a warm-up period of 5 epochs and an initial warm-up learning rate of $10^{-6}$. We utilize the cosine learning rate scheduler~\citep{DBLP:conf/iclr/LoshchilovH17} to decay the learning rate. During training, we randomly crop images to a size of $224\times224$. During the evaluation phase, for images with a resolution less than $384\times384$, we apply a center-crop with a crop ratio of 0.875. However, for images of larger sizes, we do not perform any cropping, following previous work~\citep{DBLP:conf/cvpr/0003MWFDX22}. We do not employ the EMA weights. The stochastic depth drop rates for each model are provided in Table~\ref{tab:training_config}.

\begin{table}[htpb]
    \centering
    \footnotesize
    \resizebox{1\linewidth}{!}{
    \begin{tabular}{l | c|c }
        \toprule[1.5pt]
        dataset & \multicolumn{2}{c}{\textbf{ImageNet-1K}}\\
        \midrule[1.5pt]
        configuration & TransNeXt-Micro/Tiny/Small/Base & TransNeXt-Small/Base\\
        task & $224^2$ Pre-training & $384^2$ Fine-tuning\\
        \midrule
        batch size          &  1024 &  1024 \\
        base learning rate  &  1e-3  &  1e-5 \\
        learning rate scheduler & cosine & constant \\
        min learning rate   & 1e-5  & 1e-5 \\
        training epochs     &  300  & 5   \\
        warm-up epochs      &  5    & None \\
        warm-up schedule    & linear & None  \\
        warm-up learning rate & 1e-6 & None \\
        optimizer          & AdamW & AdamW \\
        optimizer momentum & $\beta_1,\beta_2=0.9,0.999$  & $\beta_1,\beta_2=0.9,0.999$\\
        \midrule
        color jitter factor & 0.4 & 0.4\\
        auto-aug          & rand-m9-mstd0.5-inc1   & rand-m9-mstd0.5-inc1 \\
        random-erasing prob. & 0.25 & 0.25\\
        random-erasing mode  & pixel & pixel \\
        mixup  $\alpha$         & 0.8 & 0.8  \\
        cutmix $\alpha$         & 1.0 & None   \\
        mixup prob.             & 1.0 & 1.0  \\
        mixup switch prob.      & 0.5  & 0.5 \\
        \midrule
        stochastic drop path rate & 0.15/0.25/0.45/0.6 & 0.7/0.8 \\
        label smoothing        & 0.1 & 0.1  \\
        gradient clip          & 1.0 & 1.0 \\
        weight decay           & 0.05 & 0.05 \\
        exp. mov. avg. (EMA) & None & None\\
        
    \bottomrule[1.5pt]
    \end{tabular}
    }
    \vspace{-3mm}
    \caption{The pre-training and fine-tuning settings of TransNeXt on ImageNet-1K~\citep{DBLP:conf/cvpr/DengDSLL009}.}
    \label{tab:training_config}
    \vspace{-3mm}
\end{table}

\subsection{Training Settings for Downstream Tasks} 
For experiments on the ADE20K~\citep{DBLP:journals/ijcv/ZhouZPXFBT19} and COCO~\citep{DBLP:conf/eccv/LinMBHPRDZ14} datasets, we followed the training settings of Swin~\citep{DBLP:conf/iccv/LiuL00W0LG21}. We utilized the MMDetection~\citep{DBLP:journals/corr/abs-1906-07155} and MMSegmentation~\citep{mmseg2020} toolboxes for training.

For the COCO 2017 dataset~\citep{DBLP:conf/eccv/LinMBHPRDZ14}, we configured the learning rate to $10^{-4}$ and the weight decay to 0.05. In the context of the Mask R-CNN and DINO methods, the stochastic depth drop rates for TransNeXt-Tiny, TransNeXt-Small, and TransNeXt-Base were set to 0.3, 0.5, and 0.6, respectively. The model was trained for 12 epochs with a batch size of 16 using the standard 1$\times$ schedule.

For the ADE20K dataset~\citep{DBLP:journals/ijcv/ZhouZPXFBT19}, in the UperNet method, we set the learning rate to $6 \times 10^{-5}$ and the weight decay to 0.05. The stochastic depth drop rates for TransNeXt-Tiny, TransNeXt-Small, and TransNeXt-Base were set to 0.4, 0.6, and 0.7, respectively. For the Mask2Former method, we set the learning rate to $10^{-4}$ and the weight decay to 0.05, with the stochastic depth drop rates for TransNeXt-Tiny, TransNeXt-Small, and TransNeXt-Base set to 0.3, 0.5, and 0.6 respectively. All models were trained for 160K iterations with a batch size of 16 on the ADE20K dataset.

%% file: sec/appendix/ablation.tex
\section{Ablation Study}\label{sec:ablation_study}
\subsection{More Discussion on roadmap from PVT to TransNeXt}\label{sec:detailed_roadmap_PVT}

\textbf{Understanding of query embedding}: The query embedding exhibits very unique properties. Incorporating query embedding effectively improved the performance on ImageNet-1K val and ImageNet-V2 test sets but somewhat reduced performance on ImageNet-A, ImageNet-R, ImageNet-Sketch test sets; its impact on ImageNet-C was very weak. Notably, ImageNet-1K val, ImageNet-V2, and ImageNet-C (a distorted test set of ImageNet-1K val) adopted the same sampling strategy as the ImageNet-1K training set, while ImageNet-A, ImagetNet-R, and ImageNet-Sketch did not follow this principle. We believe these experimental results reflect that query embedding restricts the model's response range to enhance current task performance rather than affecting generalization to all types of data. During the learning process, the model optimizes this learnable query token, implicitly learning what the optimal question for the current task is in each attention layer (from a Visual Question Answering (VQA) perspective). This perspective can well explain why in these out-of-distribution test sets, query embedding has a very weak impact on the performance of the ImageNet-C test set which uses the same sampling strategy as the training set. Therefore, we believe there is a potential trade-off here. In the case of TransNeXt, even with query embedding, our model still achieved state-of-the-art model robustness.

\textbf{Impact of model structure}: We adjusted the width and depth of PVTv2 and the number of attention heads to match those of TransNeXt-Micro in steps 1 to 3 to avoid the impact of model structure. During this period, we observed that a deeper and thinner model significantly enhances performance. Reducing the head dimension from 48 to 24 resulted in only a 0.04\% performance change, indicating that the performance gain from increasing attention heads is extremely limited.

\subsection{Detailed Data of Multi-scale Inference}\label{sec:detailed_multiscale_inference}

\begin{table}[htpb]
\begin{center}
  \resizebox{1\linewidth}{!}{
  \begin{tabular}{l|l|ccccccc}
    \toprule[1.5pt]
     \multirow{2}{*}{Model} &\multirow{2}{*}{Method} & \multicolumn{7}{c}{Inference Size}\\
     & &$224^2$&$256^2$&$320^2$&$384^2$&$480^2$&$512^2$&$640^2$\\
    \midrule[1.5pt]
    \multirow{4}{*}{TransNeXt-Tiny} & \textbf{Normal Mode}& 84.0&84.3&84.3&84.6&83.8&83.2&\textbf{81.6}\\ 
    & No Length-scaling&84.0&84.3&84.4&84.7&83.7&83.2&80.9 \\ \
    & Interpolate RPE&84.0&84.1&84.2&84.3&83.1&82.4&79.5 \\
    & \textbf{Linear Mode}&84.0&84.0&83.9&84.1&83.0&82.6&80.7 \\
    \midrule
    \multicolumn{2}{l|}{RepLKNet-31B~\citep{DBLP:conf/cvpr/Ding0HD22}}&83.5&83.6&81.0&70.0&21.4&10.1&0.9\\
    \multicolumn{2}{l|}{SLaK-S~\citep{DBLP:conf/iclr/LiuCCCXWKPMW23}}&83.8&83.8&83.2&79.6&65.7&63.7&61.4\\  
    \multicolumn{2}{l|}{ConvNeXt-B~\citep{DBLP:conf/cvpr/0003MWFDX22}}&83.8&84.2&84.0&83.6&81.6&80.7&77.3\\
    \midrule[1.5pt]
    \multirow{2}{*}{TransNeXt-Mirco} & Normal Mode& 82.5&82.8&82.9&83.1&82.1&81.6&79.3\\ 
    & Linear Mode&82.5&82.5&82.4&82.3&80.9&80.3&77.6 \\
    \midrule[1.5pt]
    \multirow{2}{*}{TransNeXt-Small} & Normal Mode&84.7&84.9&84.9&85.0&84.1&83.8&82.2\\ 
    & Linear Mode&84.7&84.7&84.7&84.9&84.0&83.6&81.7 \\
    \midrule[1.5pt]
    \multirow{2}{*}{TransNeXt-Base} & Normal Mode&84.8&85.1&85.1&85.5&84.7&84.3&82.8\\ 
    & Linear Mode&84.8&85.0&84.9&85.1&84.1&83.5&81.5 \\
    \bottomrule[1.5pt]
    \end{tabular}
    }
\end{center}
\vspace{-3mm}
\caption{The table shows the top-1 accuracy of ImageNet-1K of $224^2$-size trained TransNeXt under \textbf{normal} and \textbf{linear} inference modes on multiple image input sizes. At the same time, the effects of length-scaled cosine attention and log-CPB on multi-scale inference were tested, and the pure convolution model was included for comparison.}
\label{table:multi_scale_inference}
\vspace{-3mm}
\end{table}

\textbf{Linear complexity mode for inference}: We observe that in Equations~\ref{eq:pfa_complexity} and ~\ref{eq:aa_complexity}, 
if we consistently set $H_p$ and $W_p$ as fixed values independent of the input size, the computational complexity of both pixel-focused attention and aggregated attention grows linearly with the length of the input sequence. In this scenario, both pixel-focused attention and aggregated attention can operate under a linear complexity mode. This linear mode endows TransNeXt with a computational complexity growth curve close to that of a pure convolutional network when inferring large-size images. We test the performance changes of $224^2$-size trained TransNeXt and two prevalent pure convolutional models at multiple resolutions. In the default normal mode, $H_p$ and $W_p$ of aggregated attention are $\frac{1}{32}$ of the input image size, while in the linear mode, $H_p$ and $W_p$ are fixed at $\frac{1}{32}$ of the training image size, \textit{i.e.}, $7\times7$.

\textbf{Results and analysis}: As shown in Table~\ref{table:multi_scale_inference} and Fig~\ref{fig:multi_scale_inference},
our TransNeXt-Tiny achieves better multi-scale extrapolation performance than pure convolutional models in both normal and linear modes. At the maximum resolution of $640^2$, the linear mode produces a performance decay of 0.5\% to 1.7\% relative to the normal mode, but such a trade-off still has advantages over pure convolutional models. As the image size increases, the performance decay of ConvNeXt-B is greater than that of TransNeXt’s linear mode. RepLKNet-31B shows a more exaggerated performance decay, with a top-1 accuracy of only 0.9\% at a resolution of $640^2$, which to some extent reveals the limitations of the super-large convolution kernel scheme. In traditional opinions, pure convolutional models have better multi-scale applicability than ViT models, and such experimental results also imply that this opinion needs to be re-examined.

\textbf{Impact of length-scaled cosine attention}: We compare the performance of length-scaled cosine attention with regular scaled cosine attention during multi-scale inference. According to Fig~\ref{fig:multi_scale_inference},
length-scaling begins to take effect when the resolution reaches $640^2$. This implies that when the sequence length variation in softmax exceeds $8\times$, longer sequence lengths begin to significantly reduce the confidence of scaled cosine attention.
 
\textbf{Extrapolation vs Interpolation for relative position bias}: When a TransNeXt model trained at a resolution of $224^2$ infers at other sizes, we default to using log-CPB~\citep{DBLP:conf/cvpr/Liu0LYXWN000WG22} to extrapolate the $B_{(i,j)\sim\sigma(X)}$ under new resolutions from spatial relative coordinates $\Delta_{(i,j)\sim\sigma(X)}$. However, generating $\Delta_{(i,j)\sim\sigma(X)}$ cannot achieve the same speed as model inference. This is not a major issue in general because when the model needs to continuously infer at one or several new sizes, we only need to pre-calculate these new $\Delta_{(i,j)\sim\sigma(X)}$ and cache them. However, when the new inference resolution of the model is unknown and needs to change instantly according to input size, we need to use traditional interpolation schemes for relative position bias to interpolate $B_{(i,j)\sim\sigma(X)}$. As depicted in Fig~\ref{fig:multi_scale_inference},
the input resolution of $640^2$ results in a significant performance degradation due to interpolation for relative position bias, surpassing that of the linear mode. This underscores the efficacy of log-CPB in extrapolating position bias. In our evaluation of UperNet with multi-scale and flip augmentations (Table~\ref{tab:ade20k_upernet}), we present test results under both interpolation and extrapolation for a balanced comparison, highlighting the influence of different schemes on multi-scale performance.

\subsection{Ablation on Positional Encoding}\label{sec:ablation_positonal_encoding}

\begin{table}[htpb]
\begin{center}
  \resizebox{1\linewidth}{!}{
  \begin{tabular}{l|c|c|c}
    \toprule[1.5pt]
     Method & Params(M) &FLOPs(G)& Top-1(\%) \\
    \midrule
    Remove $B_{(i,j)}$& 28.1 &5.6& 83.2\\
    Calculate $B_{(i,j)\sim\rho(i,j)}$ by $\textbf{log-CPB}(\Delta_{(i,j)\sim\rho(i,j)})$ & 28.2 &5.7& 83.7\\ 
    Replace $B_{(i,j)\sim\rho(i,j)}$ by $Q_{(i,j)}T$ & 28.2 &5.7& 83.4\\ 
    Replace $\textbf{log-CPB}(\Delta_{(i,j)\sim\sigma(X)})$ by learnable $B_{(i,j)\sim\sigma(X)}$ & 28.1 &5.6& \textbf{84.0}\\
    TransNeXt-Tiny & 28.2 &5.7& \textbf{84.0}\\ 
    \bottomrule[1.5pt]
    \end{tabular}
    }
\end{center}
\vspace{-3mm}
\caption{Ablation experiments on the design of relative position biases.}
\label{table:positonal_bias}
\vspace{-3mm}
\end{table}

\textbf{Results and analysis}: We conducted ablation experiments on the design of the relative position bias used in the sliding window path and pooling feature path in aggregated attention, with results shown in Table~\ref{table:positonal_bias}. When we completely removed the relative position bias $B_{(i,j)}$ used in aggregated attention, the model’s performance significantly decreased by 0.8\%. This indicates that using depthwise convolution to capture positional information from zero-padding is insufficient to represent the positional relationships of global tokens. When we used log-CPB to calculate the relative position bias of the sliding window, it also resulted in a 0.3\% performance decline. This suggests that due to different feature scales, the numerical meanings of spatial coordinates $\Delta_{(i,j)\sim\sigma(X)}$ in the pooling feature path and $\Delta_{(i,j)\sim\rho(i,j)}$ in sliding window path are not exactly the same, highlighting the importance of using different methods to learn relative position bias in the two paths. Another consideration is to use dynamic relative position bias $Q_{(i,j)}T$ calculated by positional attention to replace $B_{(i,j)\sim\rho(i,j)}$, but this resulted in a significant performance decline of 0.6\%. We believe this is due to inconsistencies in the behavior of the sliding window path and pooling path. The $\textbf{log-CPB}(\Delta_{(i,j)\sim\sigma(X)})$ calculated in the pooling path is static, while $Q_{(i,j)}T$ dynamically changes with input, and the two paths are coupled in the same softmax, causing interference with the mechanism of QKV attention.  If we also use a learnable relative position bias $B_{(i,j)\sim\sigma(X)}$ instead of calculating by $\textbf{log-CPB}(\Delta_{(i,j)\sim\sigma(X)})$ in the pooling path, it does not affect model performance, but it does cause the model to lose its ability to extrapolate position biases for unknown size inputs. This demonstrates the similarity between the relative position biases calculated through log-CPB and those directly learned, also indicating that the log-CPB module is not the source of TransNeXt’s high performance.

\subsection{Ablation on the Design of Convolutional GLU}\label{sec:ablation_convglu}

We conducted ablation experiments on the design of convolutional GLU on the CIFAR-100 dataset using a 2M-sized model. We designed three optional variants, all using GELU as the activation function:
\begin{equation}
\begin{aligned}
\textbf{ConvGLU}(X)=&(XW_1+B_1)\odot\\&\mathrm{GELU}(\mathrm{DWConv}(XW_2+B_2))
\end{aligned}
\label{eq:convglu}
\end{equation}

\begin{equation}
\begin{aligned}
\textbf{Type-1}(X)=&(XW_1+B_1)\odot\\&\mathrm{DWConv}(\mathrm{GELU}(XW_2+B_2))
\end{aligned}
\label{eq:type-1}
\end{equation}

\begin{equation}
\begin{aligned}
\textbf{Type-2}(X)=&\mathrm{DWConv}(XW_1+B_1)\odot\\&\mathrm{GELU}(XW_2+B_2)
\end{aligned}
\label{eq:type-2}
\end{equation}

\begin{equation}
\begin{aligned}
\textbf{Type-3}(X)=&\mathrm{DWConv}((XW_1+B_1)\odot\\&\mathrm{GELU}(XW_2+B_2))
\end{aligned}
\label{eq:type-3}
\end{equation}

The experiments, presented in Table~\ref{table:ablation_convglu}, showed that our convolutional GLU, which follows the design philosophy of gated channel attention, is the optimal design. In Type-1, placing DWConv after the gated activation function disrupts the effect of setting value to zero in the gating branch. In Type-2, moving DWConv to the value branch causes a significant performance drop of 0.7\% when a gating branch with a smaller receptive field controls a value branch with a larger receptive field, indicating that it is more reasonable to make gating decisions using a branch with a larger receptive field. In Type-3, adding a DWConv after the element-wise dot product result in the GLU module leads to the worst performance, suggesting that merely adding a DWConv to enhance local perceptual ability is not key to improving model performance with convolutional GLU.

\begin{table}[htpb]
\begin{center}
  \resizebox{1\linewidth}{!}{
  \begin{tabular}{l|c|c|c}
    \toprule[1.5pt]
     Design & Params. (M) & FLOPs (G) & Top-1(\%) \\
    \midrule
    ConvGLU& 2.3 & 0.5 & \textbf{82.9}\\
    Type-1& 2.3 & 0.5 & 82.6\\
    Type-2& 2.3 & 0.5 & 82.2\\
    Type-3& 2.3& 0.5 & 82.1\\
    \bottomrule[1.5pt]
    \end{tabular}
    }
\end{center}
\vspace{-3mm}
\caption{Ablation study on the design of Convolutional GLU on CIFAR-100~\citep{krizhevsky2009learning} dataset.}
\label{table:ablation_convglu}
\vspace{-3mm}
\end{table}

\subsection{Ablation on Window Size}\label{sec:ablation_window_size}

We conducted fast ablation experiments on CIFAR-100~\citep{krizhevsky2009learning} using a 2M-sized model, results are reported in Table~\ref{table:window_size}. Our observations indicate that an increase in the window size does not necessarily lead to an enhancement in the model’s performance. We believe that these experimental results are due to the introduction of pooling features provides coarse-grained global perception abilities, greatly reducing the demand for single queries to perceive the sliding window field. Moreover, the fine-grained tokens overlap with the coarse-grained tokens, leading to additional inductive bias. Since the similarity results between queries and both fine-grained and coarse-grained tokens compete in the same softmax, this approach benefits information aggregation in overlapping regions. However, as the window size increases, this inductive bias may not always be beneficial.

\begin{table}[htpb]
\begin{center}
  \resizebox{1\linewidth}{!}{
  \begin{tabular}{c|c|c|c}
    \toprule[1.5pt]
     Window Size & Params. (M) & FLOPs (G) & Top-1(\%) \\
    \midrule
    $3\times3$ & 2.3 & 0.50 & 82.9\\
    $5\times5$ & 2.3 & 0.52 & 82.0\\
    $7\times7$ & 2.3 & 0.54 & 82.9\\
    $9\times9$ & 2.3 & 0.57 & 82.5\\
    \bottomrule[1.5pt]
    \end{tabular}
    }
\end{center}
\vspace{-3mm}
\caption{The ablation results of window size. We utilized a 2M-sized TransNeXt model to conduct experiments on the CIFAR-100~\citep{krizhevsky2009learning} dataset under various window size settings. }
\label{table:window_size}
\vspace{-3mm}
\end{table}

\subsection{Ablation on Model Architecture}\label{sec:ablation_architecture}

To further explore the impact of model architecture on performance, we conducted ablation experiments based on TransNeXt-Micro. We attempted to replace aggregated attention with multi-head self-attention in stages 1-3 to observe its impact on model performance. The experimental results are presented in Table~\ref{table:ablation_architecture}. We observed that when we replaced aggregated attention with multi-head self-attention in stage 3, where the number of blocks is the highest, the model performance decreased by 0.5\%. Further replacement in stage 2 led to an additional 0.1\% decline in performance. This suggests that our aggregated attention information aggregation method has advantages over global self-attention. When we tried to replace aggregated attention in stage 1, the model encountered an out-of-memory error on $8\times$ A100s with 80GB of memory, making it impossible to train the model with this configuration.

Under a resolution of $224^2$, $7\times7$ is the smallest size that can be achieved by integer multiple downsampling. For this reason, and to maintain consistency with PVTv2, our model opted for a pooling size of $\frac{1}{32}$ at each stage. However, in stage 4, the input resolution has already been reduced to $\frac{1}{32}$, rendering the downsampling module of aggregated attention ineffective. If aggregated attention is forcibly applied at this stage, features in the sliding window would be input into softmax twice through the pooling path, leading to distortion in importance calculation. Consequently, we selected MHSA in stage 4. At larger resolutions, such as $256^2$, we can employ a pooling size of $\frac{1}{64}$ at each stage to implement a model that fully utilizes aggregated attention at all stages. As demonstrated in Table~\ref{table:ablation_architecture}, a micro-sized model that fully employs aggregated attention achieved an ImageNet-1K accuracy of 82.6\% at a resolution of $256^2$.

\begin{table}[htpb]
\begin{center}
  \resizebox{1\linewidth}{!}{
  \begin{tabular}{c|c|c|c|c|c|c}
    \toprule[1.5pt]
     Token mixer & Input size& Window size& Pool size&Params. (M) & FLOPs (G) & Top-1(\%) \\
    \midrule
    \textbf{A-A-A-M}&$224^2$&$3\times3$&$7\times7$&12.8 & 2.7 & \textbf{82.5}\\
    \textbf{A-A-M-M}&$224^2$&$3\times3$&$7\times7$&12.2 & 2.7 & 82.0\\
    \textbf{A-M-M-M}&$224^2$ &$3\times3$&$7\times7$&12.2 & 2.9 & 81.9\\
    \textbf{M-M-M-M}&$224^2$&$3\times3$&$7\times7$&12.2& 4.7 & \textbf{OOM}\\
    \midrule
    \textbf{A-A-A-A}&$256^2$&$3\times3$&$4\times4$&13.1& 3.3 & \textbf{82.6}\\
    \bottomrule[1.5pt]
    \end{tabular}
    }
\end{center}
\vspace{-5mm}
\caption{Ablation study on model architecture on ImageNet-1K dataset. \textbf{OOM} means out of memory error.}
\label{table:ablation_architecture}
\vspace{-5mm}
\end{table}

\subsection{CUDA Implementation}\label{sec:CUDA_implementation}

\begin{table*}[htpb]
\begin{center}
  \resizebox{1\linewidth}{!}{
  \begin{tabular}{l||ccc||ccc||ccc}
    \toprule[1.5pt]
     \multirow{2}{*}{Model} &\multicolumn{3}{c||}{Throughput of inference}& \multicolumn{3}{c||}{Duration of training (sec/iter)} &\multicolumn{3}{c}{Memory usage (GB)} \\
     &CUDA&Pytorch&Acceleration&CUDA&Pytorch&Acceleration&CUDA&Pytorch&Saving\\
    \midrule[1.5pt]
    TransNeXt-Micro& 1117 & 701 &+59.3\%& 0.218&0.401&+83.9\%&14.8&17.8&16.8\%\\
    TransNeXt-Tiny& 756 & 471&+60.5\% & 0.315&0.609&+93.3\%&23.2&27.3&15.0\%\\
    TransNeXt-Small& 394 & 246&+60.2\% & 0.595&1.161&+95.1\%&41.6&49.3&15.6\%\\
    TransNeXt-Base& 297 & 186&+59.6\% & 0.771 &1.568&+103.4\%&58.1&68.6&15.3\%\\
    \bottomrule[1.5pt]
    \end{tabular}
    }
\end{center}
\vspace{-5mm}
\caption{Performance comparison between CUDA implementation and native PyTorch implementation. We measure throughput using a batch size of 64 on a single V100 with 16GB of memory under FP16, while the iteration time and memory consumption during training are measured on $8\times$ A100s (PCIe) with a total batch size of 1024 under automatic mixed precision. }
\label{table:CUDA_implementation}
\vspace{-5mm}
\end{table*}

In the native PyTorch~\citep{DBLP:conf/nips/PaszkeGMLBCKLGA19} implementation, feature extraction in the sliding window path is achieved through the unfold operation. The unfold operation involves two stages: 1) extracting the tensor within the sliding window through index access, and 2) explicitly creating a large tensor copy for the extracted tensor. This explicit feature extraction operation generates a huge temporary tensor and induces memory access pressure, which significantly reduces the model’s speed. To address this, we introduce a CUDA operator implementation for calculating QK similarity and aggregating value by attention weights in the sliding window path. This implementation circumvents the need for explicit tensor extraction from the sliding window, thereby markedly enhancing the model’s throughput and training speed. As shown in Table~\ref{table:CUDA_implementation}, our CUDA implementation provides up to 60.5\% acceleration for inference, up to 103.4\% acceleration for training and saves up to 16.8\% of memory consumption for training.

On the other hand, our current implementation is solely based on native CUDA, without the introduction of substantial optimization. Consequently, it does not match the efficiency of highly optimized dense GPU operators. We evaluated the throughput of our model on a V100 16G with FP32 precision at a batch size of 64, following the methodology of Swin~\citep{DBLP:conf/iccv/LiuL00W0LG21}. As shown in Table~\ref{tab:throughput_comparsion}, our biomimetic vision implementation significantly outperforms the Focal Transformer~\citep{DBLP:journals/corr/abs-2107-00641} method in terms of both efficiency and accuracy. Moreover, under similar throughput conditions, our model demonstrates competitive top-1 accuracy on ImageNet-1K compared to previous state-of-the-art models such as MaxViT~\citep{DBLP:conf/eccv/TuTZYMBL22}, BiFormer~\citep{DBLP:journals/corr/abs-2303-08810}, and QuadTree~\citep{DBLP:conf/iclr/TangZZT22}. Despite the current throughput of the model still having a certain gap with models~\citep{DBLP:conf/cvpr/0003MWFDX22,DBLP:conf/iccv/LiuL00W0LG21} benefiting from dense operator implementations, the performance of TransNeXt is anticipated to improve with further engineering efforts. We are committed to providing more efficient operator optimizations in the future to enhance the competitiveness of TransNeXt.

\begin{table}[htpb]
\begin{center}
  \resizebox{1\linewidth}{!}{
    \begin{tabular}{l|cccc}
    \toprule[1.5pt]
    Model & \makecell{\#Params. \\ (M)} & \makecell{FLOPs \\ (G)} & \makecell{Top-1\\(\%)}& \makecell{Throughput\\(img/s)}\\ 
    \midrule[1.5pt]
    
    Swin-T~\citep{DBLP:conf/iccv/LiuL00W0LG21} & 28.3 & 4.5 & 81.2 & 790 \\ 
    ConvNeXt-T~\citep{DBLP:conf/cvpr/0003MWFDX22} & 28.6 & 4.5 & 82.3 & 779 \\ 
    MaxViT-Tiny~\citep{DBLP:conf/eccv/TuTZYMBL22} & 30.9 & 5.6 & 83.4 & 459 \\ 
    \rowcolor{LightCyan}
    \textbf{TransNeXt-Tiny (Ours)} & 28.2 & 5.7 & \textbf{84.0} & 413 \\ 
    BiFormer-S~\citep{DBLP:journals/corr/abs-2303-08810} & 25.5 & 4.5 & 83.8 & 396 \\ 
    QuadTree-B-b2~\citep{DBLP:conf/iclr/TangZZT22} & 24.2 & 4.5 & 82.7 & 361 \\ 
    Focal-T~\citep{DBLP:journals/corr/abs-2107-00641} & 29.1 & 4.9 & 82.2 & 337 \\ 
     \bottomrule[1.5pt]
    \end{tabular}
    }
    \end{center}
    \vspace{-5mm}
    \caption{Comparison of throughput between TransNeXt and its main competitor models. The throughput results were tested on a single V100 GPU under FP32 precision with a batch size of 64. The results are sorted in descending order of throughput.}
    \label{tab:throughput_comparsion}
    \vspace{-3mm}
\end{table}

%% file: sec/appendix/downstream.tex
\clearpage
\section{Downstream Experimental Results}

\begin{table}[htpb]
\begin{center}
  \resizebox{1\linewidth}{!}{
    \begin{tabular}{l|ccccccccc}
    \toprule[1.5pt]
    Backbone & \makecell{Encoder \\ size(M)} & \makecell{\#Params. \\ (M)} & \makecell{$AP^b$} & \makecell{$AP^b_{50}$}  & \makecell{$AP^b_{75}$} & \makecell{$AP^m$} & \makecell{$AP^m_{50}$} & \makecell{$AP^m_{75}$}\\
    \midrule[1.5pt]
    Swin-T~\citep{DBLP:conf/iccv/LiuL00W0LG21} & 28.3 &47.8& 43.7 &66.6&47.7& 39.8 &63.3&42.7\\
    PVTv2-B2~\citep{DBLP:journals/corr/abs-2106-13797} & 25.4 &45.3& 45.3 & 67.1&49.6&41.2&64.2&44.4 \\
    FocalNet-T (LRF)~\citep{DBLP:conf/nips/YangLDG22} & 28.6&48.9 & 46.1 & 68.2 & 50.6&41.5&65.1&44.5 \\
    Swin-S~\citep{DBLP:conf/iccv/LiuL00W0LG21} & 49.6&69.1 & 46.5 &68.7&51.3& 42.1&65.8&45.2 \\
    CSWin-T~\citep{DBLP:conf/cvpr/DongBCZYYCG22}&23&42&46.7&68.6&51.3&42.2&65.6&45.4\\
    Swin-B~\citep{DBLP:conf/iccv/LiuL00W0LG21} & 87.8&107.1 & 46.9&69.2&51.6 & 42.3&66.0&45.5 \\
    PVTv2-B3~\citep{DBLP:journals/corr/abs-2106-13797} & 45.2&64.9 & 47.0&68.1&51.7 & 42.5&65.7&45.7 \\
    InternImage-T~\citep{DBLP:conf/cvpr/WangDCHLZHLLLWQ23}&30&49&47.2&69.0&52.1&42.5&66.1&45.8\\
    PVTv2-B5~\citep{DBLP:journals/corr/abs-2106-13797} & 82.0 &101.6& 47.4 &68.6&51.9& 42.5&65.7&46.0 \\
    PVTv2-B4~\citep{DBLP:journals/corr/abs-2106-13797} & 62.6 &82.2& 47.5&68.7&52.0 & 42.7 &66.1&46.1\\
    InternImage-S~\citep{DBLP:conf/cvpr/WangDCHLZHLLLWQ23}&50&69&47.8&69.8&52.8&43.3&67.1&46.7\\
    SMT-S~\citep{DBLP:journals/corr/abs-2307-08579}&20.5&40.0&47.8&69.5&52.1&43.0&66.6&46.1\\
    BiFormer-S~\citep{DBLP:journals/corr/abs-2303-08810} & 25.5&45.2 & 47.8&69.8&52.3 & 43.2 &66.8&46.5\\
    CSWin-S~\citep{DBLP:conf/cvpr/DongBCZYYCG22}&35&54&47.9&70.1&52.6&43.2&67.1&46.2\\
    FocalNet-S (LRF)~\citep{DBLP:conf/nips/YangLDG22} & 50.3&72.3 & 48.3 &70.5&53.1& 43.1&67.4 &46.2\\
    BiFormer-B~\citep{DBLP:journals/corr/abs-2303-08810} & 56.8 &76.3& 48.6&70.5&53.8  & 43.7 &67.6&47.1\\
    CSWin-B~\citep{DBLP:conf/cvpr/DongBCZYYCG22}&78&97&48.7&70.4&53.9&43.9&67.8&47.3\\
    InternImage-B~\citep{DBLP:conf/cvpr/WangDCHLZHLLLWQ23}&97&115&48.8&70.9&54.0&44.0&67.8&47.4\\
    SMT-B~\citep{DBLP:journals/corr/abs-2307-08579}&32&51.7&49.0&70.2&53.7&44.0&67.6&47.4\\
    FocalNet-B (LRF)~\citep{DBLP:conf/nips/YangLDG22} & 88.7 &111.4 & 49.0 &70.9&53.9& 43.5&67.9&46.7 \\
    \rowcolor{LightCyan}
    \textbf{TransNeXt-Tiny (Ours)} &28.2& 47.9 & \textbf{49.9} &\textbf{71.5} &\textbf{54.9} & \textbf{44.6} & \textbf{68.6}& \textbf{48.1}\\ 
    \rowcolor{LightCyan}
    \textbf{TransNeXt-Small (Ours)} & 49.7& 69.3 & \textbf{51.1}&\textbf{72.6}&\textbf{56.2}&\textbf{45.5}&\textbf{69.8}&\textbf{49.1}\\ 
    \rowcolor{LightCyan}
    \textbf{TransNeXt-Base (Ours)} &89.7& 109.2&\textbf{51.7}&\textbf{73.2}&\textbf{56.9}&\textbf{45.9}&\textbf{70.5}&\textbf{49.7}\\ 
     \bottomrule[1.5pt]
    \end{tabular}
    }
    \end{center}
    \vspace{-5mm}
    \caption{Detailed COCO object detection and instance segmentation results using the Mask R-CNN~\citep{DBLP:journals/pami/HeGDG20} $1\times$ schedule, sorted in ascending order based on $AP^b$ performance.}
    \label{tab:coco_maskrcnn}
    \vspace{-5mm}
\end{table}

\begin{table}[htpb]
\begin{center}
  \resizebox{1\linewidth}{!}{
    \begin{tabular}{l|ccccccc}
    \toprule[1.5pt]
     Model & \makecell{Encoder \\ size(M)} &\makecell{\#Params.\\(M)}&Epochs& \makecell{scales}& \makecell{Pre-trained} &\makecell{$AP^b$}\\
     \midrule[1.5pt]
         ConvNeXt-B~\citep{DBLP:conf/cvpr/0003MWFDX22}&88.6&110&12&4&IN-1K ($384^2$)&52.6\\
         ConvNeXt-L~\citep{DBLP:conf/cvpr/0003MWFDX22}&198&221&12&4&IN-1K ($384^2$)&53.4\\
         \rowcolor{LightCyan}
         \textbf{TransNeXt-Tiny (Ours)}&\textbf{28.2}&\textbf{47.8}&12&4&IN-1K (\textbf{224}$^2$)&\textbf{55.1}\\
         \rowcolor{LightCyan}
         \textbf{TransNeXt-Tiny (Ours)}&28.2&48.1&12&5&IN-1K ($224^2$)&\textbf{55.7}\\
         \rowcolor{LightCyan}
         \textbf{TransNeXt-Small (Ours)}&49.7&69.6&12&5&IN-1K ($224^2$)&\textbf{56.6}\\
         \rowcolor{LightCyan}
         \textbf{TransNeXt-Base (Ours)}&89.7&110&12&5&IN-1K ($224^2$)&\textbf{57.1}\\
         \midrule
        Swin-L~\citep{DBLP:conf/iccv/LiuL00W0LG21}&197&218 &12&5&IN-22K ($384^2$)&57.2\\
     \bottomrule[1.5pt]
    \end{tabular}
    }
    \end{center}
    \vspace{-5mm}
    \caption{Comparison of object detection results on the COCO dataset using the DINO method. The data for Swin~\citep{DBLP:conf/iccv/LiuL00W0LG21} is sourced from MMDetection~\citep{DBLP:journals/corr/abs-1906-07155}, while the data for ConvNeXt~\citep{DBLP:conf/cvpr/0003MWFDX22} is referenced from detrex~\cite{ren2023detrex} project. The results are sorted in ascending order based on the $AP^b$ scores.}
    \label{tab:coco_dino}
    \vspace{-5mm}
\end{table}

\begin{table}[htpb]
\begin{center}
  \resizebox{1\linewidth}{!}{
    \begin{tabular}{l|cccccc}
    \toprule[1.5pt]
     Model & \makecell{Encoder \\ size(M)} & \makecell{\#Params. \\ (M)} & \makecell{Crop\\-size}& \makecell{Pre-trained} &\makecell{mIoU\\(\%)} & \makecell{ +MS\\(\%)} \\
     \midrule[1.5pt]
        Swin-T~\citep{DBLP:conf/iccv/LiuL00W0LG21}&28.3&60&$512^2$&IN-1K&44.5&45.8\\
        Focal-T~\citep{DBLP:journals/corr/abs-2107-00641}&29.1&62&$512^2$&IN-1K&45.8&47.0\\
        ConvNeXt-T~\citep{DBLP:conf/cvpr/0003MWFDX22}&28.6&60&$512^2$&IN-1K&46.0&46.7\\
        FocalNet-T(LRF)~\citep{DBLP:conf/nips/YangLDG22}&28.6&61&$512^2$&IN-1K&46.8&47.8\\
        Swin-S~\citep{DBLP:conf/iccv/LiuL00W0LG21}&49.6&81&$512^2$&IN-1K&47.6&49.5\\
        UniFormer-S~\citep{DBLP:conf/iclr/Li00S00022}&22&52&$512^2$&IN-1K&47.6&48.5\\
        Focal-S~\citep{DBLP:journals/corr/abs-2107-00641}&51.1&85&$512^2$&IN-1K&48.0&50.0\\
        Swin-B~\citep{DBLP:conf/iccv/LiuL00W0LG21}&87.8&121&$512^2$&IN-1K&48.1&49.7\\
        
        ConvNeXt-S~\citep{DBLP:conf/cvpr/0003MWFDX22}&50.2&82&$512^2$&IN-1K&48.7&49.6\\
        Focal-B~\citep{DBLP:journals/corr/abs-2107-00641}&89.8&126&$512^2$&IN-1K&49.0&50.5\\
        FocalNet-S(LRF)~\citep{DBLP:conf/nips/YangLDG22}&50.3&84&$512^2$&IN-1K&49.1&50.1\\
        ConvNeXt-B~\citep{DBLP:conf/cvpr/0003MWFDX22}&88.6&122&$512^2$&IN-1K&49.1&49.9\\
        SMT-S~\citep{DBLP:journals/corr/abs-2307-08579}&20.5&50.1&$512^2$&IN-1K&49.2&50.2\\
        SMT-B~\citep{DBLP:journals/corr/abs-2307-08579}&32&61.8&$512^2$&IN-1K&49.6&50.6\\
        UniFormer-B~\citep{DBLP:conf/iclr/Li00S00022}&49.8&80&$512^2$&IN-1K&50.0&50.8\\
        FocalNet-B(LRF)~\citep{DBLP:conf/nips/YangLDG22}&88.7&126&$512^2$&IN-1K&50.5&51.4\\

        \rowcolor{LightCyan}
        \textbf{TransNeXt-Tiny (Ours)}&28.2&59&$512^2$&IN-1K&\textbf{51.1}&\textbf{51.5/51.7}\\
        \rowcolor{LightCyan}
        \textbf{TransNeXt-Small (Ours)}&49.7&80&$512^2$&IN-1K&\textbf{52.2}&\textbf{52.5/52.8}\\
        
        ConvNeXt-B~\citep{DBLP:conf/cvpr/0003MWFDX22}&88.6&122&$640^2$&IN-22K&52.6&53.1\\
        \rowcolor{LightCyan}
        \textbf{TransNeXt-Base (Ours)}&89.7&121&\textbf{512}$^2$&\textbf{IN-1K}&\textbf{53.0}&\textbf{53.5/53.7}\\
     \bottomrule[1.5pt]
    \end{tabular}
    }
    \end{center}
    \vspace{-5mm}
    \caption{A comprehensive comparison of semantic segmentation results on the ADE20K dataset using the UperNet method. +MS for evaluation with multi-scale and flip augmentations. In the context of multi-scale evaluation, TransNeXt reports test results under two distinct scenarios: interpolation and extrapolation of relative position bias. The results are sorted in ascending order based on the mIoU scores.}
    \label{tab:ade20k_upernet}
    \vspace{-5mm}
\end{table}

\begin{table}[htpb]
\begin{center}
  \resizebox{1\linewidth}{!}{
    \begin{tabular}{l|ccccccc}
    \toprule[1.5pt]
     Model & \makecell{Encoder \\ size(M)} &\makecell{\#Params.\\(M)}&\makecell{Crop\\-size}& \makecell{Pre-trained} &\makecell{mIoU(\%)}\\
     \midrule[1.5pt]
         Swin-S~\citep{DBLP:conf/iccv/LiuL00W0LG21}&49.6&68.8&$512^2$&IN-1K ($224^2$)&51.2\\
         Swin-B~\citep{DBLP:conf/iccv/LiuL00W0LG21}&87.8&107&$640^2$&IN-1K ($224^2$)&52.4\\
         \rowcolor{LightCyan}
         \textbf{TransNeXt-Tiny (Ours)}&\textbf{28.2}&\textbf{47.5}&\textbf{512}$^2$&IN-1K ($224^2$)&\textbf{53.4}\\ 
         Swin-B~\citep{DBLP:conf/iccv/LiuL00W0LG21}&87.8&107&$640^2$&IN-22K ($384^2$)&53.9\\
         \rowcolor{LightCyan}
         \textbf{TransNeXt-Small (Ours)}&\textbf{49.7}&\textbf{69.0}&\textbf{512}$^2$&\textbf{IN-1K (224}$^2$\textbf{)}&\textbf{54.1}\\
         \rowcolor{LightCyan}
         \textbf{TransNeXt-Base (Ours)}&89.7&109&\textbf{512}$^2$&\textbf{IN-1K (224}$^2$\textbf{)}&\textbf{54.7}\\
     \bottomrule[1.5pt]
    \end{tabular}
    }
    \end{center}
    \vspace{-5mm}
    \caption{Comparison of semantic segmentation results on the ADE20K dataset using the Mask2Former method. The data for Swin~\citep{DBLP:conf/iccv/LiuL00W0LG21} is sourced from MMSegmentation~\citep{mmseg2020}. The results are sorted in ascending order based on the mIoU scores.}
    \label{tab:ade20k_mask2former}
    \vspace{-5mm}
\end{table}

%% file: sec/appendix/visualization.tex
\section{Visualization Based on Effective Receptive Field}

\begin{figure*}[t]
\centering
    \includegraphics[width=0.64\textwidth]{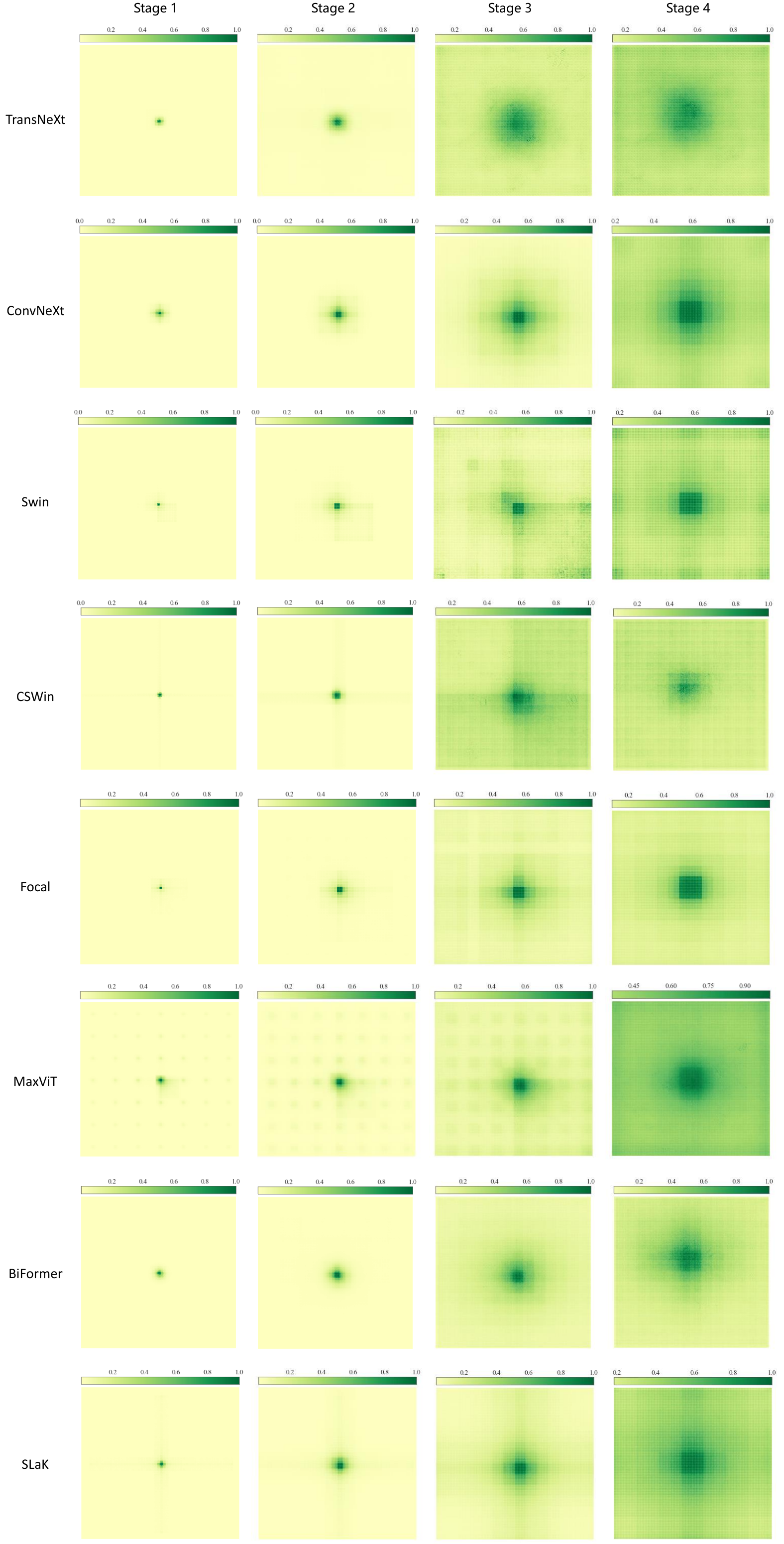}
    \vspace{-2mm}
  \caption{Visualization of the Effective Receptive Field (ERF) on ImageNet-1K validation set. Each visualization is based on an average of 5000 images with a resolution of $224\times224$. We visualize the ERFs of four stages for eight models: TransNeXt-Tiny, ConvNeXt-T, Swin-T, CSWin-T, Focal-T, MaxViT-Tiny, BiFormer-S, and SLaK-T.}
  \label{fig:erf_stage}
  \vspace{-7mm}
\end{figure*}

\begin{figure*}[t]
\centering
    \includegraphics[width=0.65\textwidth]{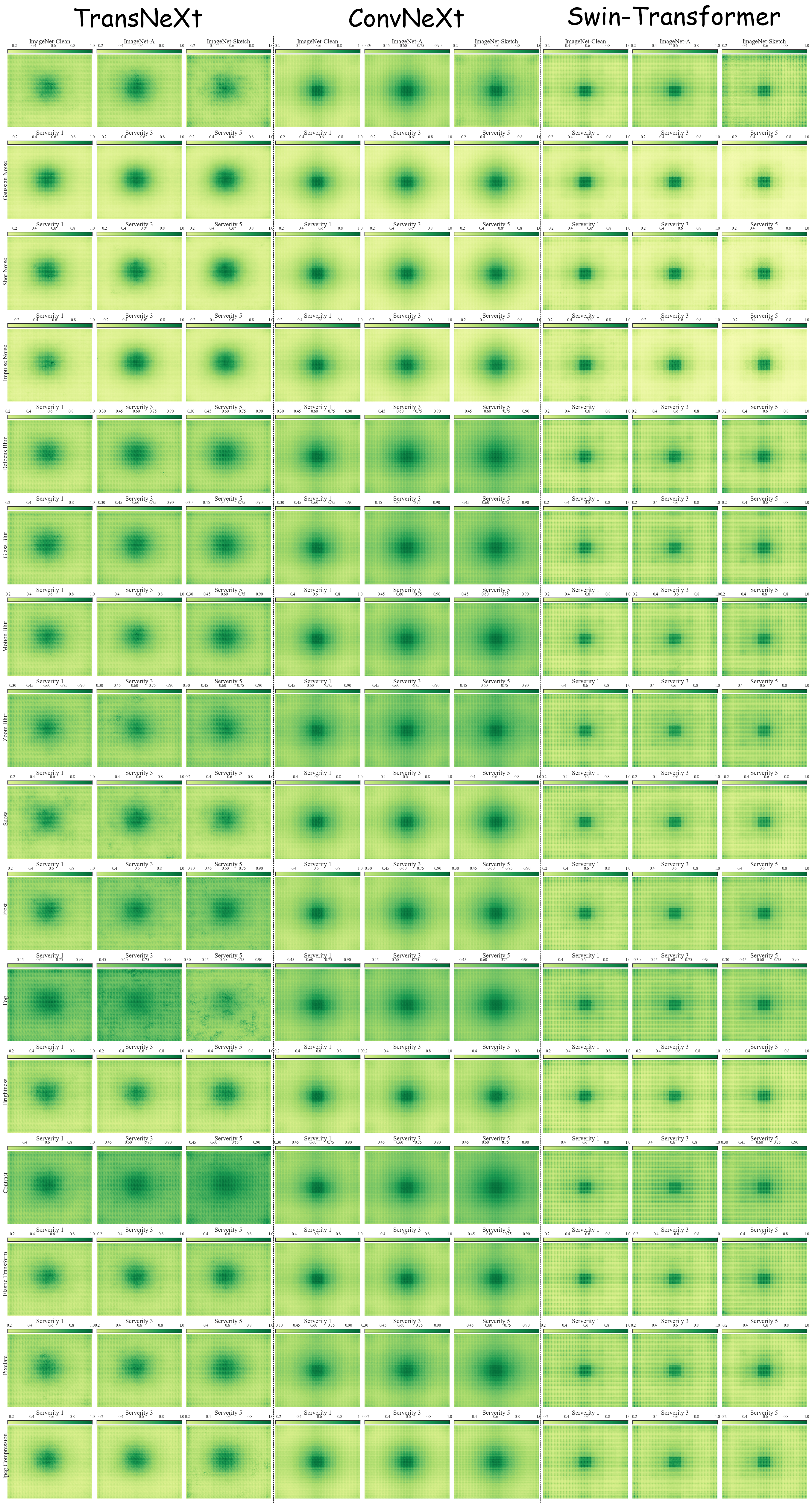}
  \caption{Visualization of the Effective Receptive Field (ERF) for TransNeXt-Tiny, ConvNeXt-T, and Swin-T on various datasets including ImageNet-1K validation set (Clean), ImageNet-Adversarial, ImageNet-Sketch, and ImageNet-C. The visual analysis diagrams for ImageNet-C commence from the second row of the figure. For each corruption mode, we have included visual images with severity levels of 1, 3, and 5. Each ERF image is generated by averaging over 5000 images with a resolution of $224 \times 224$ from each dataset.}
  \label{fig:erf_robust}
\end{figure*}

We employ the Effective Receptive Field (ERF)~\citep{DBLP:conf/nips/LuoLUZ16} method as a visualization tool to analyze the information aggregation approach of TransNeXt. In Fig~\ref{fig:erf_stage}, we visualize the ERF of four encoder stages for eight models: TransNeXt-Tiny, ConvNeXt-T, Swin-T, CSWin-T, Focal-T, MaxViT-Tiny, BiFormer-S, and SLaK-T. In Fig~\ref{fig:erf_robust}, we further conduct a comprehensive ERF visualization comparison on the fourth stage of the models on ImageNet-A, ImageNet-Sketch, and ImageNet-C datasets.

Our observations are as follows:
\begin{enumerate}[leftmargin=*]
\item As depicted in Fig~\ref{fig:erf_stage}, in the comparative visualization of ERF for multi-stage outputs, TransNeXt-Tiny outperforms seven other models in ERF coverage at the third stage, exhibiting a more natural and smoother visual perception. This can partially explain TransNeXt’s performance advantage in detection and segmentation tasks, as these tasks rely more heavily on lower-stage outputs.

In contrast, ConvNeXt, Swin, and CSWin exhibit distinct blocky patterns, which we attribute to artifacts from their token mixer designs. Despite the presence of multiple layers, these token mixers are unable to eliminate artifacts induced by window-based local attention or convolution kernels, resulting in an unnatural information mixing. 

Notably, although Focal Transformer has designed a unique biomimetic attention mechanism, it is not exempt from this issue. The method it employs, which is based on window partitioning, still results in significant blocky artifacts. These blocky artifacts are also markedly evident in MaxViT, which utilizes a hybrid CNN-ViT architecture. Moreover, the grid sampling methodology employed in MaxViT’s token mixer introduces additional grid artifacts, further demonstrating the prevalence of these artifact phenomena across different models and designs.

Cutting-edge ViT and convolutional models have made some improvements in this regard. BiFormer utilizes a data-driven approach that enables window-based ViT models to autonomously select window combinations, thereby circumventing the unnatural traces caused by manual window design. SLaK employs an ultra-large convolution kernel scheme to achieve global perception, thus mitigating the receptive field degradation caused by depth degradation in convolutional models. We observe that these methods have to some extent alleviated the unnatural visual perception caused by manual window design or convolution kernel stacking in previous ViT and convolutional models, but the unnatural blocky artifacts caused by window partitioning and convolution kernel stacking are still not entirely eliminated.

This observation supports the experimental evidence that deep networks with residual blocks function as ensembles of shallower networks, highlighting the significance of a single token mixer in achieving a local-global modeling approach that is more akin to biological vision. TransNeXt’s ERF represents an information perception methodology that aligns more closely with biological vision, achieving a natural visual perception and validating the effectiveness of its biomimetic design.

\item As shown in Fig~\ref{fig:erf_robust}, in a comprehensive visualization evaluation across multiple out-of-distribution test sets, TransNeXt-Tiny demonstrates a more adaptive information perception method. Its effective receptive field’s information perception method undergoes significant changes with different datasets. This change can be clearly observed at multiple severity levels on ImageNet-C. Meanwhile, Swin-T’s ERF exhibits a similar pattern across all test sets, and ConvNeXt-T’s ERF lies somewhere in between. We believe that a more adaptive ERF reflects the model’s robustness, and such visualization comparison results are consistent with the robustness evaluation results in Table~\ref{tab:image_classification_robustness}.
\end{enumerate}

